%% file: iclr2022_conference.tex
\documentclass{article} 
\usepackage{iclr2022_conference,times}

\input{math_commands.tex}

\usepackage{hyperref}
\usepackage{url}
\usepackage{booktabs}       
\usepackage{amsfonts}       
\usepackage{nicefrac}       
\usepackage{microtype}      
\usepackage{xcolor}         

\usepackage{graphicx}
\usepackage{multirow}
\usepackage{subcaption}
\usepackage{amsmath}
\usepackage{amsfonts}
\usepackage{bm}
\usepackage{pifont}
\usepackage{array}
\usepackage{diagbox}

\usepackage{cancel}
\usepackage{soul}
\usepackage{mathtools}

\usepackage{bbm}
\usepackage{xfrac}
\usepackage{color}

\usepackage{wrapfig}

\usepackage{enumitem}

\usepackage{multicol}
\usepackage{nicematrix}
\usepackage{hhline}
\usepackage{threeparttable}
\usepackage{makecell}
\usepackage{tabularx}

\newcommand{\papertitle}{PIM-QAT: Neural Network Quantization for Processing-In-Memory (PIM) Systems}

\title{\papertitle}

\input{tex_files/author}

\newcommand{\cmark}{\ding{51}}%
\newcommand{\xmark}{\ding{55}}%

\newcommand{\etal}{\emph{et al}.}

\newcommand{\jqk}[1]{\textcolor{black}{#1}}
\newcommand{\zyc}[1]{\textcolor{black}{#1}}
\newcommand{\zycii}[1]{\textcolor{black}{#1}} 

\newcommand{\IMC}[1][]{\ensuremath{\mathrm{PIM}}}
\newcommand{\supp}{Appendix}

\newcommand{\round}[1][]{\ensuremath{\mathrm{round}}}
\newcommand{\qeq}[1][]{\ensuremath{\stackrel{\scalebox{1.01}{\textbf{?}}}{=}}}

\renewcommand{\E}[1]{\mathbb{E}[#1]}
\newcommand{\VAR}[1]{\mathbb{VAR}[#1]}

\definecolor{mybrown}{RGB}{237, 125, 49}
\definecolor{myblue}{RGB}{0, 176, 240}

\renewcommand{\eqref}[1]{(\ref{#1})}

\newtheorem{thm}{Theorem}
\newtheorem{assumpt}{Assumption}

\iclrfinalcopy 
\begin{document}

\maketitle

\vskip -3ex

\input{tex_files/0_abstract}
\input{tex_files/1_introduction}
\input{tex_files/2_background}

\input{tex_files/3_method}

\input{tex_files/4_experiment}

\input{tex_files/5_conclusion}

\bibliographystyle{apalike}
\bibliography{mybib, IMC}

\clearpage

\newpage

\input{tex_files/appendix_into_main}

\end{document}

%% file: math_commands.tex
\usepackage{amsmath,amsfonts,bm}

\def\eqref#1{equation~\ref{#1}}

\def\1{\bm{1}}

\DeclareMathAlphabet{\mathsfit}{\encodingdefault}{\sfdefault}{m}{sl}
\SetMathAlphabet{\mathsfit}{bold}{\encodingdefault}{\sfdefault}{bx}{n}

\newcommand{\E}{\mathbb{E}}



%% file: tex_files/author.tex
\author{Qing Jin$^{1}$\thanks{Equal contribution}\,\;, Zhiyu Chen$^2$\footnotemark[1]\,\;, Jian Ren$^3$, Yanyu Li$^1$, {\bf Yanzhi Wang$^1$, Kaiyuan Yang$^2$} \\
$^1$Northeastern University, $^2$Rice Univeristy, $^3$Snap Inc.\\
\texttt{\{jinqingking\}@gmail.com}, \texttt{\{zc37, kyang\}@rice.edu}\\\texttt{\{jren\}@snapchat.com},\texttt{\{li.yanyu, yanz.wang\}@northeastern.edu}}

%% file: tex_files/0_abstract.tex
\begin{abstract}

Processing-in-memory (PIM), an increasingly studied neuromorphic hardware, promises orders of energy and throughput improvements for deep learning inference. Leveraging the massively parallel and efficient analog computing inside memories, PIM circumvents the bottlenecks of data movements in conventional digital hardware.~\zyc{However, an \textbf{extra quantization} step (i.e. PIM quantization), typically with limited resolution due to hardware constraints, is required to convert the analog computing results into digital domain. Meanwhile, non-ideal effects extensively exist in PIM quantization because of the imperfect analog-to-digital interface, which further compromises the inference accuracy. Due to hardware limitations, PIM systems decompose the bulky matrix multiplication into smaller subsets, making the computing flow fundamentally different from the conventionally quantized models.}~\jqk{
In this paper, we propose a method for training quantized networks to incorporate PIM quantization, which is ubiquitous to all PIM systems. Specifically, we propose a PIM quantization aware training (PIM-QAT) algorithm, and introduce rescaling techniques during backward and forward propagation by analyzing the training dynamics to facilitate training convergence. We also propose two techniques, namely batch normalization (BN) calibration and adjusted precision training, to suppress the adverse effects of non-ideal linearity and stochastic thermal noise involved in real PIM chips. Our method is validated on three mainstream PIM decomposition schemes, and \textbf{physically on a prototype chip}. Comparing with directly deploying conventionally trained quantized model on PIM systems, which does not take into account this extra quantization step and thus fails, our method provides significant improvement. It also achieves comparable inference accuracy on PIM systems as that of conventionally quantized models on digital hardware, across CIFAR10 and CIFAR100 datasets using various network depths for the most popular network topology.}
\end{abstract}

%% file: tex_files/1_introduction.tex
\section{Introduction}
\label{sec:intro}
\vspace{-4pt}
Recent progress of deep learning has witnessed great success in a wide range of applications at the cost of enormous computations and energy budget. To alleviate the resource constraints and enable deep learning inference on pervasive mobile and edge devices, extensive research has been conducted on algorithm optimizations for conventional digital hardware (e.g. GPU, CPU), with the goals of compressing models and reducing the number of operations~\citep{pact,quantefficienttraining,sat,liu2020cocopie,ye2018progressive,zhang2018systematic,dorefa,sun2020ultra,mishra2017wrpn}.
{On the other hand, hardware innovations for deep learning focus on building dedicated devices with optimized dataflow and reusing to minimize data movement~\citep{ chen2016eyeriss, chen2014dadiannao,du2015shidiannao, jouppi_-datacenter_2017}, which is the well known energy and latency bottleneck in deep learning and many other data-centric computations~\citep{horowitz20141}}.


\input{figs/fig_pim_quant}

\textbf{\emph{Processing in-memory}} (PIM), {inspired by neuromorphic engineering, attracts increasing attention as a potential hardware solution to data movement bottlenecks}~\citep{ambrogio2018equivalent, ielmini2018memory, jia_programmable_2020, prezioso_training_2015, xue2020cmos,yao2020fully, zhang2017memory}. By performing computations directly inside the weight storage memories, PIM promises significantly reduced data traffic between the memory and computing units. The merits of PIM over conventional digital hardware are three folds. {First, the data movement energy and latency can be alleviated. Second, massively parallel computing, like multiply-and-accumulate (MAC), in memory arrays greatly amortize total energy and area. Third, the computation in memory is essentially in analog, which is known to be more efficient than in digital for low-precision computation. As an example, a recent PIM demonstration \citep{yao2020fully} achieves 110$\times$ higher energy efficiency and 30$\times$ better compute density than TESLA V100 GPU. Meanwhile, PIM systems can be built upon various types of integrated memory technologies, from static random-access memory (SRAM) that scales well with Moore's law, to emerging non-volatile memories that 
stores an analog weight in a tiny unit, e.g. resistive random-access memory (ReRAM)~\citep{prezioso_training_2015, xue2020cmos, yao2020fully} and phase change memory (PCM)~\citep{ambrogio2018equivalent,joshi2020accurate}. }




\zyc{Despite the forthcoming efficiency and throughput gains, PIM systems require an extra quantization step to digitize the analog MAC results because the high-precision scaling multiplication is more efficient in digital domain (see Fig.~\ref{fig:pim_quant}). However, such extra quantization typically has limited resolution (typically 5-8 bit) due to the hardware constraints and thus leads to significant inference accuracy loss. Moreover, as shown in the right of Fig.~\ref{fig:pim_quant}, conventional digital quantization is more flexible to quantize in a very small sub-range of the whole output by scaling, clipping and rescaling before bit-truncating, which effectively achieves an arbitrarily small LSB. On the contrary, PIM quantization involved in modern PIM systems~\citep{ams, biswas2019conv, jia2021scalable, lee2021fully, lee2021charge} typically only supports direct bit-truncating, mainly because accurate scaling operations in analog domain will lead to unaffordable energy and area overhead that is potentially even larger than the whole PIM system~\citep{lee2021charge}. This direct bit-truncating introduces significant information loss~\citep{ams}, which makes PIM quantization drastically different and more challenging than the digital counterparts.
}Furthermore, the inevitable \textbf{non-idealilies} in the PIM quantization, including the imperfect linearity and random thermal noise of the analog-to-digital converters (ADCs), aggravate the side-effect of the low-resolution quantization and turn the conventionally quantized model into random guess, as shown in Fig.~\ref{fig:framework}. Limited by the memory array size and analog computing precision, as well as to reduce the input range for less quantization errors, PIM systems compute MACs in a $k$-bit-serial fashion ($1\leq k\leq$ input/weight bit-width) and decompose the channels into multiple subsets. The partial sums of PIM output are then re-combined via digital shift-and-adds or accumulation (see Fig.~\ref{fig:pim_quant}). As the computing flow is fundamentally different from conventional models, a new method specialized for PIM systems, taking the \textbf{decomposition}, \textbf{quantization}, as well as \textbf{recombination} into account, is highly desired.


In this paper, we systematically analyze the discrepancies between PIM and conventional digital hardware, and propose \emph{PIM quantization aware training} (\textbf{PIM-QAT}) for deep neural networks. Note that in this work we focus on the extra quantization step involved in all types of PIM systems, as mentioned above, and only consider imperfect linearity and stochastic thermal noise. 
More sophisticated cases of hard-to-model non-linearities caused by inaccurate storage of weights or other effects like data retention issues are out of scope of this work, as they are less general but specific to some types of PIM systems, such as ReRAM.
Our method is ideally suitable for the SRAM PIM, where only non-idealities coming from ADCs play a role. However, the problem of PIM quantization is general enough and ubiquitous to all other types of PIM systems, which share the same computing flow as ours, despite their different memory technologies and hardware topologies, including PCM and ReRAM PIMs. Therefore, our method is general and will greatly benefit models running on these systems.
We summarize our contributions as the following:

\begin{itemize}[noitemsep,nolistsep,leftmargin=3ex]
\item We propose PIM-QAT based on a basic assumption of \textit{generalized straight-through estimator} (\textbf{GSTE}). GSTE is a generalization of the famous straight-through estimator (STE)~\citep{ste}, which has been adopted in conventional quantization~\citep{dorefa}.
\item We study the \textit{training dynamics} unique to the PIM-QAT, and propose \textit{scaling techniques} for both forward and backward propagation during training to tackle convergence problems.
\item We leverage \textit{Batch Normalization (BN) calibration} to close the gap between idealized training and real-case inference on real PIM systems with fabrication and run-time variations.
\item We further propose an \textit{adjusted precision training} algorithm and study the potential relations between training precision and the effective number of bits (ENOB) of the actual physical PIM system for inference.
\item We test the proposed method on three major PIM decomposition schemes (native, bit serial, differential) that cover the majority of PIM hardware designs. We extensively evaluate the method on a \textbf{silicon prototype} of SRAM PIM with realistic non-idealities. A micrograph of the prototype chip is shown in Fig.~\ref{fig:framework}.
\end{itemize}


\input{figs/fig_framework}

%% file: figs/fig_pim_quant.tex
\begin{figure*}[t]
\centering
\hspace*{-3ex}
\begin{subfigure}[t]{.3\linewidth}
\centering
\includegraphics[width=.8\linewidth]{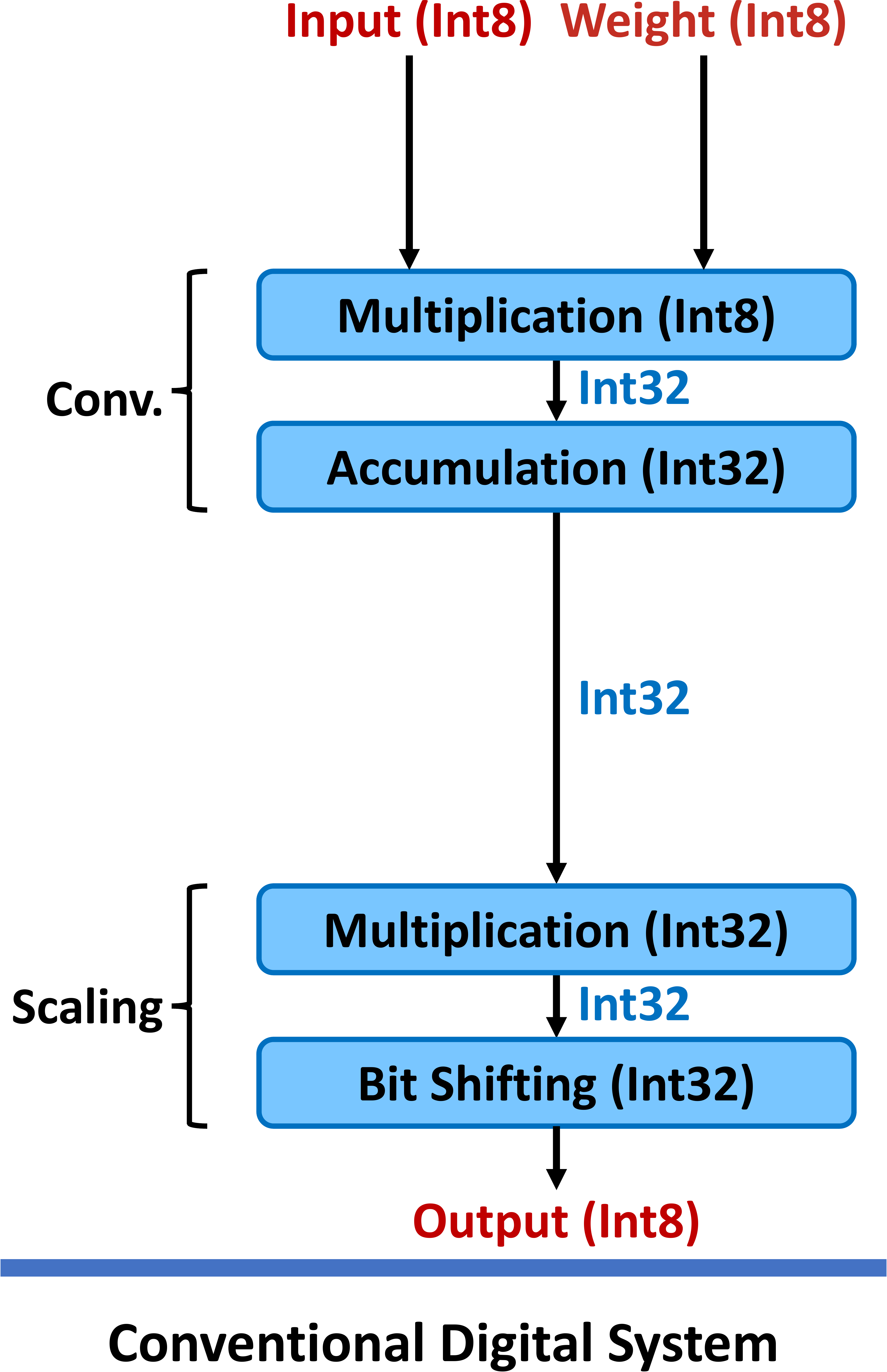}
\end{subfigure}
\hskip -3ex
\begin{subfigure}[t]{.375\linewidth}
\raisebox{0.00\linewidth}{
\centering
\includegraphics[width=.8\linewidth]{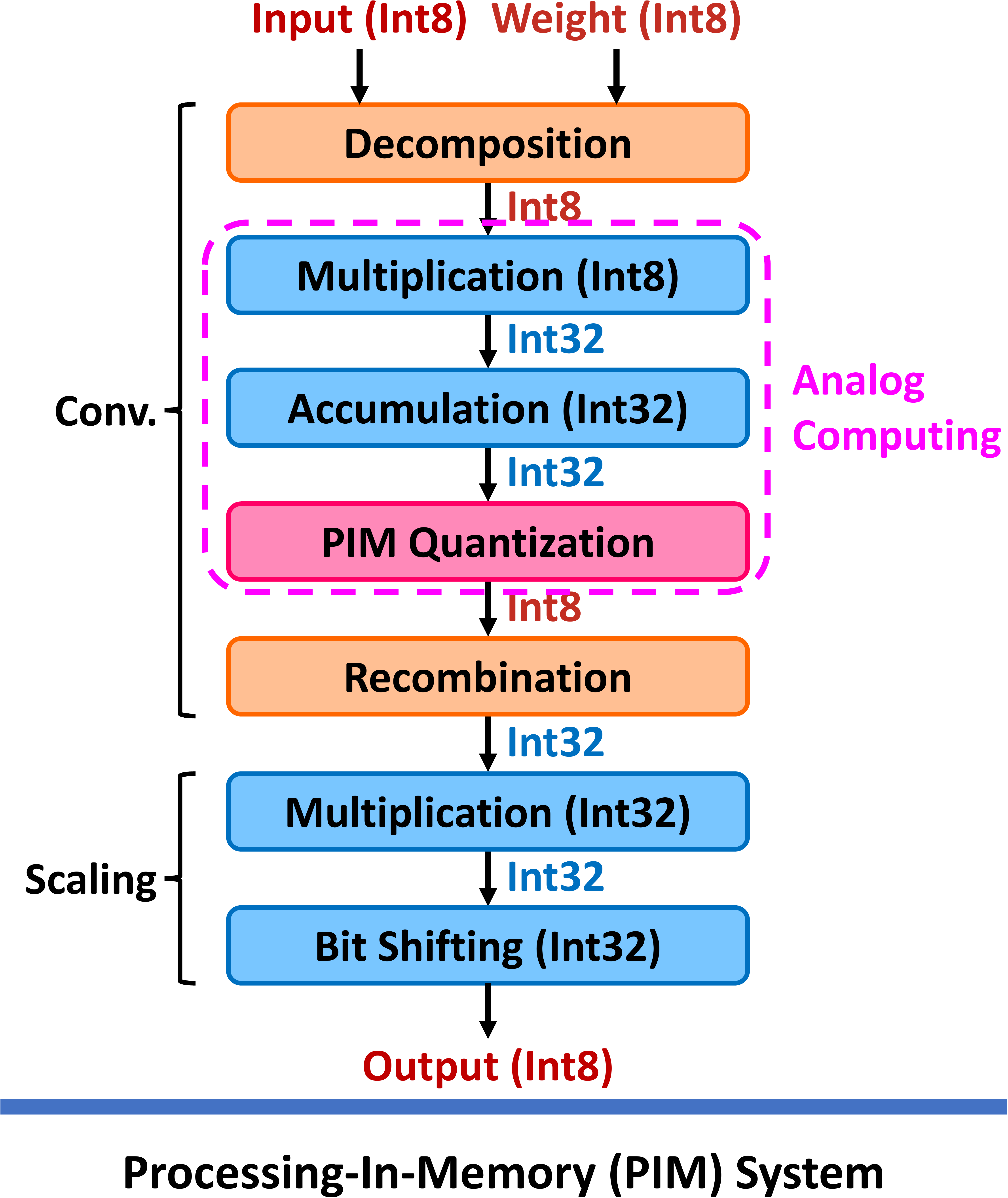}
}
\end{subfigure}
\hskip -8ex
\begin{subfigure}[t]{.42\linewidth}
\raisebox{0.12\linewidth}{
\centering
\includegraphics[width=.9\linewidth]{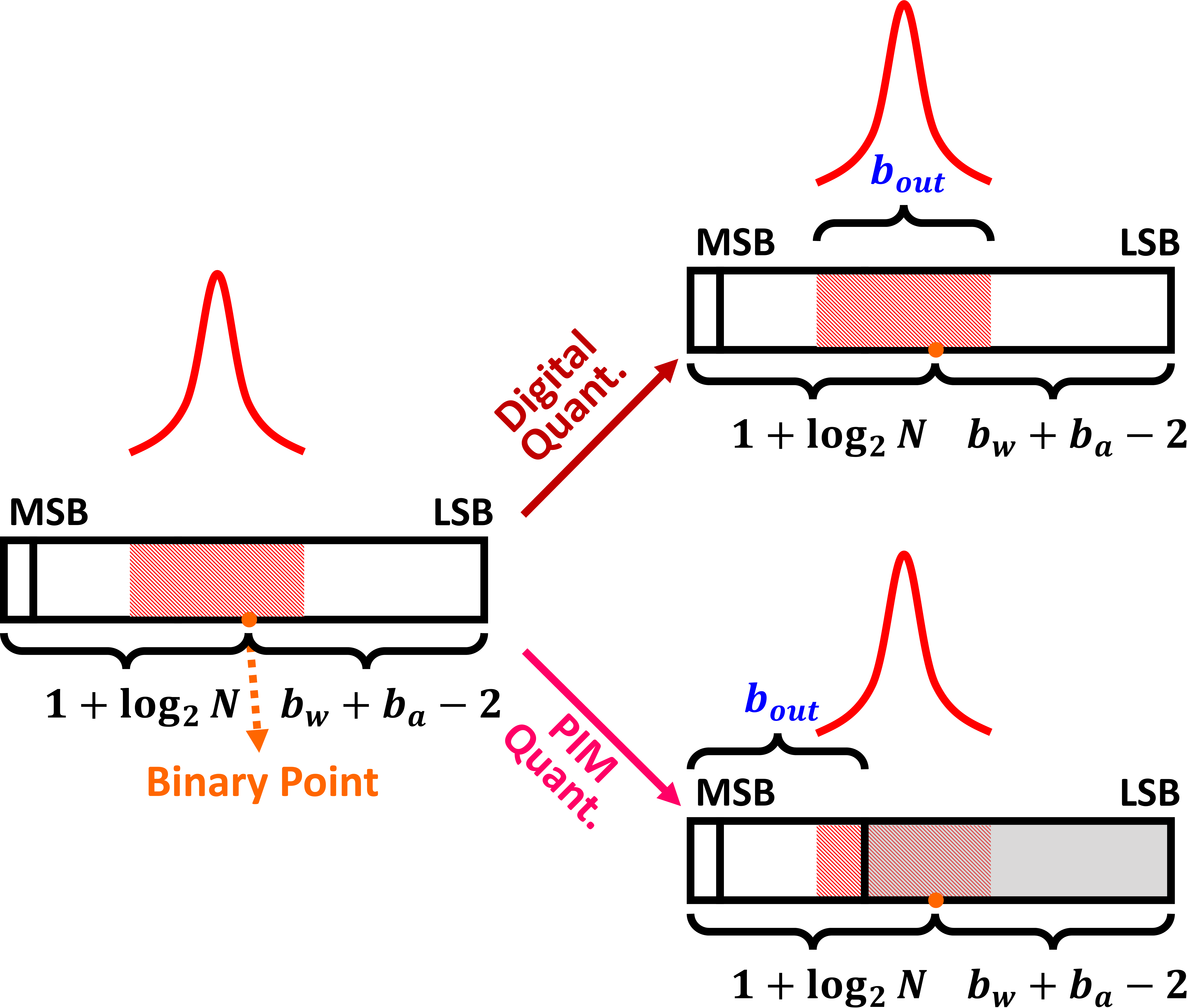}
}
\end{subfigure}
\caption{
Comparison between conventional digital systems (left) and the processing-in-memory (PIM) systems (middle), and their quantization effect (right). Unlike conventional digital systems where quantization is only applied once on both inputs and weights for efficient integer convolution, processing-in-memory (PIM) systems require an \textbf{extra quantization} due to \textbf{limited resolution} of the analog-to-digital interface.
As shown in the right, unlike conventional digital quantization that is flexible to quantize in sub-range with an arbitrarily small LSB via scaling and clipping, PIM quantization in state-of-the-art PIM systems~\citep{ams, biswas2019conv, jia2021scalable, lee2021fully, lee2021charge} typically performs direct bit-truncating (i.e. discarding the LSBs and keeping the MSBs), mainly because accurate scaling operations in analog domain will lead to unaffordable energy and area overhead that is potentially even larger than the whole PIM system~\citep{lee2021charge}. This direct bit-truncating introduces significant information loss~\citep{ams} and severely deteriorates the accuracy performance of model running on it. PIM quantization is thus drastically different and more challenging than the digital counterparts.
Note the input to PIM quantization can have a much smaller range than 32-bit integers and here we use ``INT32'' to denote the most general case.
}
\label{fig:pim_quant}
\end{figure*}

%% file: figs/fig_framework.tex
\begin{figure*}[t!]
\centering
\includegraphics[width=0.9\linewidth]{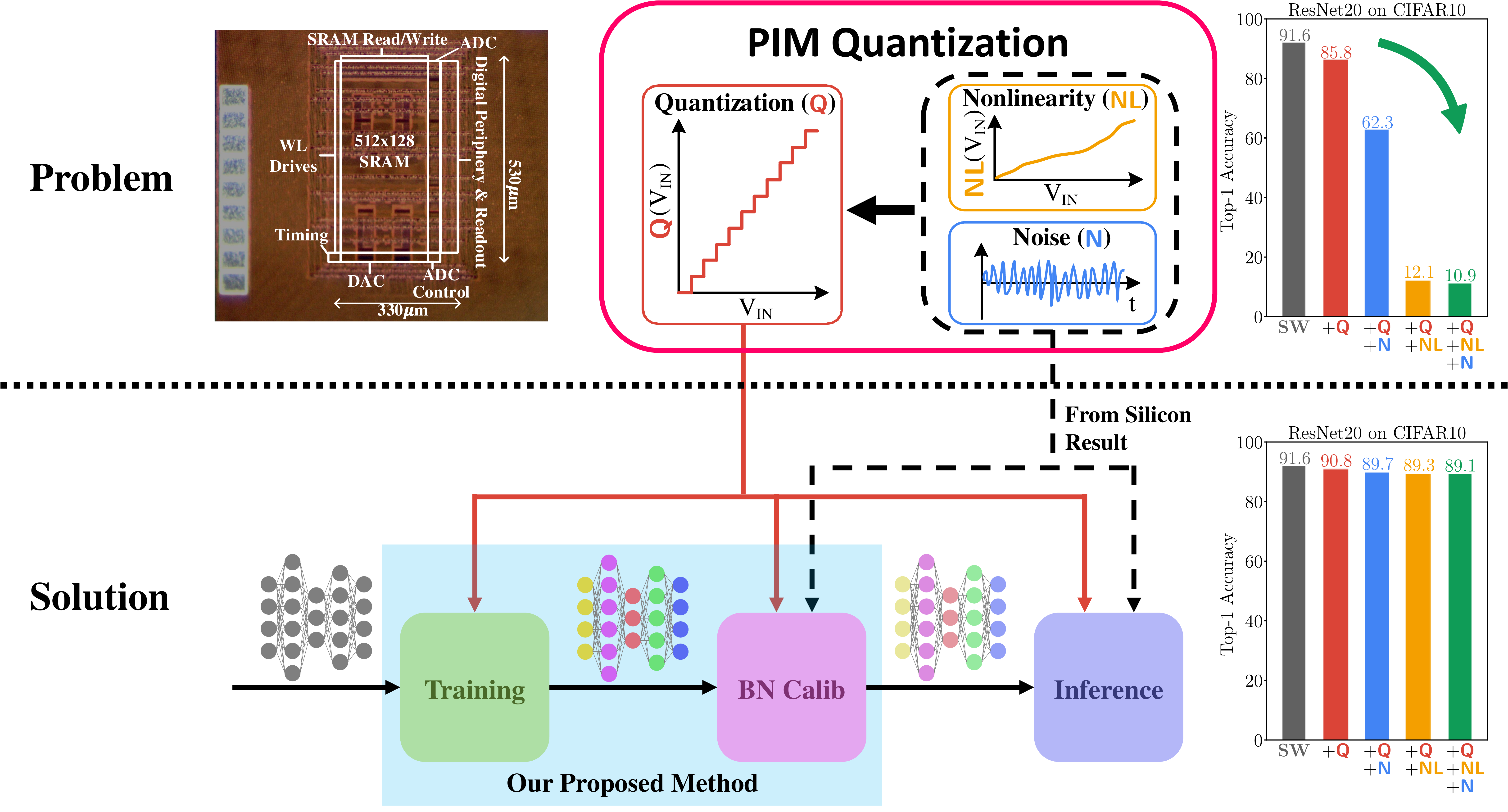}
\caption{
Imperfect MAC in a processing in-memory (PIM) system and our proposed solution (workflow). 
\jqk{Compared to conventional digital systems, the extra low-resolution quantization step in PIM systems introduces significant information loss, making models trained with conventional quantization techniques fail. The two non-idealities of this extra quantization, namely imperfect linearity (which we simply denote as non-linearity) and stochastic thermal noise, further aggravate the errors of PIM-quantization. On the contrary, our method takes into account the ideal PIM quantization during training, and applies BN calibration and adjusted precision training algorithm to alleviate the impact of the two non-ideal effects. Note that non-idealities are not directly modeled during training because PIM quantization in different chips exhibits different linearity and noise behaviors due to inter-die variations. Therefore, training with limited non-ideal samples may lead to biased results. Our techniques reduce the accuracy gap and improve the robustness for real PIM systems.}
}\label{fig:framework}
\end{figure*}

%% file: tex_files/2_background.tex
\section{Background and Related Work}

\paragraph{Processing In-Memory Hardware}
\zyc{Low-precision PIM quantization is ubiquitous in state-of-the-art PIM systems. Depending on different accuracy targets and model sizes, the quantization resolution ranges from 1-bit \citep{yin2020xnor} to 8-bit \citep{jia202115}, and most of them introduce large quantization errors. The possible levels of the analog MAC results can be up to 67.5$\times$ larger than the quantization levels \citep{lee2021fully}. On the other hand, different PIM systems adopt different decomposition strategies. The maximum number of elements ($N$) in one analog MAC is an important parameter because a larger $N$ brings more energy savings, but also extends the levels of analog MACs (which is proportional to $N$). In reality, $N$ is selected from 9 \citep{yoon202140} to 2304 \citep{valavi_64-tile_2019}, making the effect of channel-wise decomposition unique in different PIM systems. Meanwhile, weights are stored in different formats as digital memories (e.g. SRAM) only store 1-bit data in each cell while analog memories (e.g. ReRAM) have multi-state storage, and inputs are decomposed depending on the resolution of digital-to-analog converters (DACs). As a result, different memory topologies lead to different 
PIM decomposition
schemes and quantization errors. Our proposed method unifies all the design choices above and tackles the quantization challenges under various hardware settings.}


Despite the potential accuracy loss, PIM is a promising approach for deep learning applications due to its high energy efficiency. Table 1 summarizes the efficiency of V100 GPU~\citep{V100}, TPU~\citep{jouppi_-datacenter_2017}, ReRAM PIM~\citep{yao2020fully}, and our SRAM PIM prototype, which represents ``peak'' energy efficiency at 100\% utilization of the hardware. Training techniques specific for PIM systems is thus an urgent demand.


\input{tabs/tab_efficiency_comp}



\paragraph{Analog Computing/PIM Aware Quantization}
\zycii{Several prior studies~\citep{ams, he2019noise, joshi2020accurate, long2020q} improve inference accuracy by incorporating PIM non-idealities or quantization effects into training. He~\etal~\citep{he2019noise} and Joshi~\etal~\citep{joshi2020accurate} develop a noise-injection approach to tolerate the data storage errors (e.g. conductance drift, inaccurate data programming, IR drop, etc.) that exist in multi-state non-volatile memories. However, both studies fail to model the PIM quantization in a pratical way, where they either ignore the quantization step during inference ~\citep{joshi2020accurate} or assume a power-hungry analog scaling operation ~\citep{he2019noise}. Q-PIM~\citep{long2020q} simplifies the model quantization without the need of retraining, yet ignores all analog non-idealities but only supports digital PIM platforms that have limited applications.}
On the other hand, Rekhi~\etal~\citep{ams} propose a more general analog/mixed signal (AMS) error model, where PIM quantization together with its non-idealities are summarized into an additive noise determined by the effective number of bits (ENOB) of the whole system. Such a high-level abstraction is broadly applicable to different PIM decomposition schemes without considering the detailed implementations, but it also renders sub-optimal results. As shown in Table~\ref{tab:method_comp}, it is unclear how to estimate ENOB for complex PIM \zyc{decomposition} schemes such as bit serial and differential. Meanwhile, different ENOBs require individually trained models, and the underlying assumption of having a sufficiently large $N$ for central limit theorem does not hold for many practical PIM systems. In this paper, we attempt to solve this discrepancy by incorporating a more interpretable and white-box model for any given PIM hardware in the training procedure. 


\input{tabs/tab_method_comp}

%% file: tabs/tab_efficiency_comp.tex
\begin{table}[htb]
\caption{Energy efficiency of different hardware.}
\label{tab:energy_efficiency_comp}
\vspace{-8pt}
\begin{center}
\scalebox{.85}{
\begin{tabular}{ccccc}
\toprule
\multirow{2}{*}{Hardware} & V100 & \multirow{2}{*}{TPU} & {ReRAM} & SRAM \\
& GPU && PIM & (Ours) \\
\midrule
Efficiency & \multirow{2}{*}{0.1} & \multirow{2}{*}{2.3} & \multirow{2}{*}{11} & \multirow{2}{*}{49.6} \\
(TOPS/W) & & & &\\
\bottomrule
\end{tabular}
}
\end{center}
\vskip -6ex
\end{table}

%% file: tabs/tab_method_comp.tex
\begin{table}[htb]
\caption{Comparison of training methods for neural networks applied on processing in-memory (PIM) systems.}
\label{tab:method_comp}
\vspace{-8pt}
\begin{center}
\scalebox{0.85}{
\begin{tabular}{cccc}
\toprule
& Native & Bit Serial & Differential  \\
\midrule
Baseline & \xmark & \xmark & \xmark \\
AMS & \cmark & \xmark & \xmark \\
Ours & \cmark & \cmark & \cmark \\
\bottomrule
\end{tabular}
}
\end{center}
\vskip -3ex
\end{table}

%% file: tex_files/3_method.tex
\section{PIM Quantization Aware Training}

In this section, we first describe a generic model of the extra quantization involved in typical PIM systems, and introduce our basic assumption - generalized straight-through estimator (GSTE). Based on these, we propose our PIM-QAT method (Fig.~\ref{fig:framework}), including two scaling techniques to stabilize training dynamics, BN calibration to adapt to fabrication variations of PIM hardware, and an adjusted precision training approach to account for stochastic thermal noise and imperfect linearity together with its chip-to-chip variations.



\subsection{Problem Definition}
Multiply-and-accumulate (MAC) is the basic operation involved in typical neural networks, including convolution, recurrent, fully-connect, as well as attention layers. Compared to software implementation with a digital system, where the inner product of weight $W_i$ and $x_i$ is given by $y=\sum\limits_{i=1}^{N}W_ix_i$,
the output of inner product implemented on a generic PIM system can be formulated as
\begin{equation}
\label{eq:pim}
    \widetilde{y}_{\IMC}=\bm{\mathsf{Q}}(\bm{\mathsf{NL}}(\sum_{i=1}^N\widetilde{Q}_i\widetilde{q}_i);b_{\IMC})+\varepsilon
\end{equation}
Here, $\widetilde{Q}_i\in[-1,1]$ and $\widetilde{q}_i\in[0,1]$ are quantized weights and activations, with $b_w$ and $b_a$ bits, respectively. $\bm{\mathsf{Q}}$ and $\bm{\mathsf{NL}}$ denote quantization and imperfect linearity, and $\varepsilon$ is the stochastic thermal noise introduced by the system. $b_{\IMC}$ is the precision for PIM quantization $\bm{\mathsf{Q}}$. 
Eqn. \zyc{(\ref{eq:pim}) represents one MAC operation in PIM system (see Analog Computing in Fig.~\ref{fig:pim_quant}), and is generic to different PIM decomposition schemes including native, bit serial, as well as differential schemes (see Sec. 4, also see \supp~\ref{sec:proof}). Note that the variations of $\widetilde{Q}_i$ and $\widetilde{q}_i$ are not considered here as those non-idealities are only general in analog memories (e.g., ReRAM) but have minor effects in digital memories (e.g., SRAM). We leave this feature as a future investigation.}


\subsection{Generalized Straight-Through Estimator}
In order to take the full advantage of and adapt the neural network to PIM systems, we need to make training aware of $\bm{\mathsf{Q}}$. For this purpose, we first investigate the conventional quantization-aware training targeting digital accelerators.
Generally, in order to back-propagate through a quantized neural network, where the non-differentiable function $\round(\cdot)$ is extensively used, the typical practice is to adopt the straight-through estimator~\citep{ste} as proposed in~\citep{dorefa}, where for a real input $r_i\in[0,1]$, the derivative of quantized output with respect to the input is given by
\begin{equation}
\label{eq:STE}
    \frac{\partial}{\partial r_i}\Big(\frac{1}{2^k-1}\round\big((2^k-1)r_i\big)\Big)=1
\end{equation}
Here, $k$ is the number of bits for quantization.

To evaluate the effect of $\bm{\mathsf{Q}}$ involved in PIM systems for both forward and backward propagation,
we first generalize the STE result in equation~\eqref{eq:STE} to a stronger yet more flexible assumption, which we name as \textit{generalized straight-through estimator} (GSTE) and is summarized in Assumption~\ref{assumpt:gste}. 


\begin{assumpt}[Generalized STE]\label{assumpt:gste}
The differential of the round function is given by
\begin{equation}
\label{eq:GSTE}
    \mathrm{d}\,\round(x)=\xi\cdot\mathrm{d}x
\end{equation}
where $\xi$ is a scaling factor assigned empirically.
\end{assumpt}


Note that GSTE can also be viewed as a definition for the differential of the discontinuous function $\round(\cdot)$, and equation~\eqref{eq:STE} can be easily derived from it by setting $\xi=1$. In practice, $\xi$ can be set to different values for different scenarios (for example, for different bit-widths or inputs). We will elaborate more on this point in the following. GSTE will serve as the basis for our whole analysis, and as shown in the~\supp, from GSTE we can derive the following theorem for PIM-QAT.


\begin{thm}[PIM Quantization Aware Training]\label{thm:imc}
For ideal PIM systems with PIM decomposition schemes including native, bit serial, as well as differential, where the extra quantization taken into account during forward propagation is ideal without imperfect linearity or noise involved, the backward propagation takes exactly the same form as that for conventional quantization, with only the quantized quantity involved are adjusted accordingly. Specifically, for a PIM system with quantized weight $\widetilde{Q}_i\in[-1,1]$ of $b_w$ bits and quantized input $\widetilde{q}_i\in[0,1]$ of $b_a$ bits, the forward and backward propagation are given by
\begin{subequations}
\begin{align}
    \mathbf{Forward\colon}&\widetilde{y}_{\IMC}=\bm{\mathsf{Q}}\bigg(\sum_{i=1}^N\widetilde{Q}_i\widetilde{q}_i;b_{\IMC}\bigg)\\
    \mathbf{Backward\colon}&\mathrm{d}\widetilde{y}_{\IMC}=\xi\cdot\mathrm{d}\bigg(\sum_{i=1}^N\widetilde{Q}_i\widetilde{q}_i\bigg)\label{eq:imc_bwd}
\end{align}
\end{subequations}
respectively, where $N$ is the total number of MACs of the inner product 
and $b_{\IMC}$ is PIM bit-width. For conventional quantization with digital accelerator, we have $b_{\IMC}=+\infty$ and the forward propagation is reduced to the typical case of $\widetilde{y}=\sum\limits_{i=1}^N\widetilde{Q}_i\widetilde{q}_i$.
\end{thm}

Theorem~\ref{thm:imc} demonstrates that quantization introduced by PIM systems only alters the forward propagation and impacts the calculated values of outputs, but does not change the way of taking derivative over inputs and weights. Additionally, it enables awareness of such quantization during gradient calculation, which is critical for optimization of neural networks targeting PIM systems.

\subsection{Rescaling}

With Theorem~\ref{thm:imc}, we are ready to incorporate PIM quantization during training. However, 
this does not guarantee good performance, 
which also relies on a stable training determined by training dynamics~\citep{he2015delving,poole2016exponential,schoenholz2016deep,yang2017mean}. In a well-trained model, gradients from different layers should be on the same order to guarantee backward information propagation, in order to avoid gradient exploding/vanishing problems~\citep{bengio1994learning,hochreiter1991untersuchungen,hochreiter2001gradient,pascanu2013difficulty}. 
As shown in~\supp~\ref{sec:scale_enlarging}, PIM quantization has a scale-enlarging effect, especially for low bit-width. To understand the impact of this effect, we first introduce the following theorem.


\begin{thm}[Training Dynamics]\label{thm:training}
For a neural network composed of repeated blocks, where each block is a sequential of a fully-connected layer, some nonlinear effect (for example, the PIM quantization operation), an extra scaling, a batch normalization layer, and the nonlinear activation $\varphi(\cdot)$, as defined as following
\begin{subequations}
\begin{align}
    x^{(l+1)}_i &= \varphi(y^{(l)}_i)\\
    y^{(l)}_i &= \gamma^{(l)}_i\frac{z^{(l)}_i-\mu^{(l)}_i}{\sigma^{(l)}_i} + \beta^{(l)}_i\\
    z^{(l)}_i &=\eta^{(l)}\widetilde{z}^{(l)}_i\label{eq:scale_fwd}\\
    \widetilde{z}^{(l)}_i &= f(W^{(l)}_{ij},x^{(l)}_j)\sim\rho^{(l)}\sum_{j=1}^{n^{(l)}}W^{(l)}_{ij}x^{(l)}_j\label{eq:scale_nonlinear}
\end{align}
\end{subequations}
where $x^{(l)}$ is the input to the $l$-th block, $W^{(l)}$ is the weight matrix of the fully-connected layer, ${n^{(l)}}$ is the number of input neurons, $f$ represents the nonlinear effect, $\rho^{(l)}$ is introduced to demonstrate the effect of the nonlinearity on the scale of output standard deviation, $\eta^{(l)}$ is an extra scaling factor introduced and explained in the following, and $\gamma^{(l)}$, $\beta^{(l)}$, $\sigma^{(l)}$, $\mu^{(l)}$ are parameters and running statistics of the batch norm layer.
If the differential of the nonlinear effect $f$ is given by
\begin{equation}
    \mathrm{d}\widetilde{z}^{(l)}_i =\xi^{(l)}\cdot\mathrm{d}\bigg(\sum_{j=1}^{n^{(l)}}W^{(l)}_{ij}x^{(l)}_j\bigg)\label{eq:scale_bwd}
\end{equation}
where $\xi^{(l)}$ is the scaling factor for backward propagation inside the $l$-th layer, then for zeroth order approximation (mean-field assumption), the activation gradient variance ratio between two adjacent layers is given by
\begin{equation}
\begin{aligned}
    \frac{\VAR{\partial_{x^{(l)}}\mathcal{L}}}{\VAR{\partial_{x^{(l+1)}}\mathcal{L}}}\approx\left(\frac{\xi^{(l)}}{\rho^{(l)}}\right)^2\cdot\frac{n^{(l+1)}}{n^{(l)}}\label{eq:grad_ratio}
\end{aligned}
\end{equation}
\end{thm}
Theorem~\ref{thm:training} indicates that the scale ratio between activation gradients from two adjacent layers depends on the scaling factors introduced during forward and backward for the nonlinear effect.

Based on the results in~\eqref{eq:grad_ratio}, we can find that if we do not introduce extra scaling factor as in~\eqref{eq:scale_bwd} but 
follow the conventional practice of STE (in other words, $\xi^{(l)}=1$ for all $l$), 
the scale-enlarging effect
may cause gradient exploding/vanishing problem. Proper intialization as proposed in~\cite{he2015delving} is not effective in this case. 
Experiment demonstrates that for some PIM decomposition scheme (such as bit serial and differential) and sufficiently low bit-width (such as $7$-bit), the training does not converge.

To overcome this problem, we propose to scale the gradient according to~\eqref{eq:scale_bwd}, and determine the necessary scale by also calculating the standard deviation of the result from software quantization. Specifically, the scaling factor in~\eqref{eq:scale_bwd} is given by
\begin{equation}
    \xi=\sqrt{\frac{\VAR{y_{\IMC}}}{\VAR{y}}}
\end{equation}
where $y_{\IMC}$ is the result with PIM system and $y$ is that with conventional software. Note that this only introduces extra computation during training as the scale factor is only necessary for backward to stablize training, and will not impact the inference procedure. Experiment demonstrates that this backward scaling solves the problem for cases those otherwise do not give reasonable results.

Besides scaling for backward, we find that scaling during forward with predefined constant factor helps training, especially for low bit-widths, such as those lower than $5$-bit. Even for higher precision, introducing extra scaling can still be beneficial. However, as shown in equation~\eqref{eq:grad_ratio}, the ratio does not depend on this factor, as it should be absorbed into the running variance of the following batch normalization layers. We guess this is related to numerical stability for computation, but the underlying mechanism is still unclear to us and we leave it as a future work. However, we list the scaling factor that we find best for practice in the~\supp. 

\subsection{BN Calibration}





In the above we discuss about PIM systems with ideal quantization, where the PIM quantization is perfectly linear without stochastic thermal noise. For real systems, there are two non-ideal effects. First, \zyc{the circuit non-idealities in the analog-to-digital conversion will degrade the quantization linearity}. Second, random fluctuations in the circuit will add thermal noise on the quantized output. Moreover, the imperfectness accompanying the linear mapping varies from chip to chip, and there lacks a unified model to describe such variation accurately. On the other hand, direct training with injected noise can either deteriorate the training progress (for example, if the noise injected is too large), or the noise energy can be different for different real systems. Consequently, it is almost impossible to directly consider these effects during training, especially in backward propagation.

Experiments demonstrate that the non-idealities have the potential to change the BN statistics (see~\supp~\ref{sec:bn_stats_nonideality}), and following~\citep{yu2019universally}, we propose to use a small portion of training data and calibrate BN running statistics before evaluation. For both BN calibration and final inference, we apply exactly the same real-case non-idealities. We find this can significantly improve the performance, especially when the non-ideal effect is strong (e.g., more imperfect linearity or larger injected noise).~\jqk{Note that~\cite{joshi2020accurate} exploits calibrating batch normalization statistics for the purpose of accuracy retention involved in PCM systems, which is a different problem from ours.} In~\supp~\ref{sec:more_bn_calib}, we present more experiments, where we find that BN calibration is able to reduce the impact of gain and offset in PIM quantization and thus alleviates hardware calibration efforts.

\subsection{Adjusted Precision Training}

\input{figs/fig_enob_nonideality}

Besides BN calibration, we study the possibility of employing different precisions for training and inference. The reasoning behind is that the non-idealities only affect the least significant bits during the involved quantization mapping, which effectively reduces the number of distinguishable output levels from the PIM system. To quantify this reduction, ADC designs typically use a metric called the effective number of bit-width (ENOB) and it can be adopted here.
As an example, Fig.~\ref{fig:enob_nonideality} shows that 
the standard deviation of MAC computing errors in a 7-bit PIM system will be equal to that of ideal lower bit PIM systems, when random noise is added. Note that this adjusted precision training method considers both noise injection and imperfect linearity. Depending on quantization bit-widths, noise levels, imperfect linearity forms, the optimal training precision varies but is expected to be always smaller than the ideal PIM resolution.

\vskip -4ex

%% file: figs/fig_enob_nonideality.tex
\begin{figure}[t]
\vskip -2ex
\centering
\includegraphics[width=.5\linewidth]{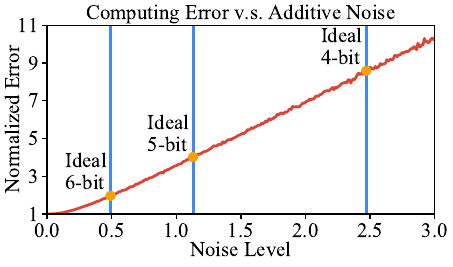}
\caption{
Computing error as a function of the standard deviation of additive noise in our 7-bit PIM chip.
}
\label{fig:enob_nonideality}
\vskip -2ex
\end{figure}

%% file: tex_files/4_experiment.tex
\section{Experiments}

\input{tabs/tab_vanilla}

\paragraph{Native Scheme.} We first investigate the possibility of directly applying the conventional quantized model on PIM system, which serves as the baseline for our comparison. To this end, we take the native scheme as an example, and fix the number of multiplications for each processing to $9$, namely we use a unit channel of $1$ to split the input channels. We experiment on CIFAR10 with ResNet20, and the results are summarized in Table~\ref{tab:vanilla}. Our method significantly outperforms baseline, especially for ultra-low bit-widths. As shown in Table~\ref{tab:vanilla}, the AMS method in~\citep{ams} is supposed to work for the native scheme. It indeed improves over the baseline but shows inferior performance than ours. These results demonstrate that PIM quantization has non-negligible impacts on the final accuracy, and it is necessary to take this quantization into account during training for optimal inference accuracy on PIM systems. 

\vspace{-1ex}
\paragraph{Real Chip Results.} 


We experiment on CIFAR10 and CIFAR100, with several ResNet models as well as one modified VGGNet11 following~\citep{jia_programmable_2020}. We also use different numbers of unit channels, namely $8$ and $16$, to split the input channels, corresponding to number of computing units of $72$ and $144$, respectively. As shown in Table~\ref{tab:cifar_results}
, our method provides significantly better results than the baseline. Specifically, prediction in the baseline models is barely better than random guess, meaning the non-idealities from the real chip corrupt the behavior of neural networks trained in this way. In contrast, our method gives comparable results as those on digital system (the software results), meaning the trained models are robust to real-case non-idealities. \zycii{Moreover, VGGNet shows less accuracy loss than ResNet because the 
more redundant model has better tolerance over the real-chip non-idealities. It is widely-used for PIM platforms with high-accuracy requirements~\citep{jia2021scalable, lee2021fully}}. Note that using smaller $N$ typically leads to better performance, especially for CIFAR100, at the cost of reduced throughput and energy efficiency.

\input{tabs/tab_cifar_results}

\vspace{-1ex}
\paragraph{Other PIM Decomposition Schemes.}


We further verify our method on three other most common PIM decomposition schemes,
including native, differential and bit serial (see~\supp~\ref{sec:proof}). We experiment on ideal PIM with different inference resolutions and noise levels. As shown in Figure~\ref{fig:noise}, we compare our method with the baseline using BN calibration on ResNet20 with CIFAR10 dataset. It is clear that for all schemes with different resolution and noise levels, our method is consistently superior, especially for high noise level and for differential and bit-serial schemes, both of which are more practical and complex than the native one. This justifies that our proposed method is applicable to a wide range of PIM implementations.

\input{figs/fig_adjusted_bit_training}
\input{figs/fig_noise}

\vspace{-1ex}

\paragraph{Adjusted Precision Training.}
Here we provide some ablation studies on adjusted precision training.
We use an ideal PIM system with bit serial scheme as the example. For different inference resolutions and noise levels, the best accuracy with the optimal training resolution is illustrated in Fig.~\ref{fig:adjusted_bit_training}, where the accuracy is directly listed and different colors denote different training precision adjustments. We find that for low noise level, it is optimal to train the model with the same resolution as that for inference, and for larger noise, 
it is better to use a smaller one due to reduced ENOB. Moreover, we find that the noise level threshold of adjusting the training resolution depends on the absolute value of inference resolution, and higher inference resolution tends to be more sensitive to 
noise and requires precision adjustment for a lower noise level threshold. There is clear correlation between the precision reduction and ENOB, but they are not exactly the same. 
This should be related to the varying sensitivity of inference on each MAC operation. More in-depth analysis of the relation between ENOB and training setting is beyond the scope of this work and left for future study.

Our analysis and experiments demonstrate that naively deploying neural network quantized with conventional method on PIM systems is problematic and ineffective, and PIM quantization has non-negligible impact on final performance. Incorporating it into training is critical and will improve the accuracy to a large extent. 
It also inspires and provides a desirable starting point for future research to incorporate hardware-specific behaviors into algorithm co-design for energy-efficient analog computing systems. Such efforts will bridge the gap between hardware and software developments to achieve unprecedented energy efficiency, while maintaining a competitive neural network performance.

\vspace{-6pt}

%% file: tabs/tab_vanilla.tex
\begin{table}[htb]
\vskip -3ex
\caption{Effect of PIM quantization on accuracy of neural networks (ResNet20 for CIFAR10) trained with different methods for native scheme ($N=9$). Baseline refers to model trained with conventional quantization method~\citep{sat}. AMS refers to model trained with the method in~\citep{ams}.}
\label{tab:vanilla}
\vspace{-8pt}
\begin{center}
\begin{small}
\setlength{\tabcolsep}{10pt}
\renewcommand{\arraystretch}{1.0}
\scalebox{0.9}{
\begin{NiceTabular}{ccc|ccc}
\toprule
$b_{\IMC}$ & Method & Acc. & $b_{\IMC}$ & Method & Acc. \\
\midrule
\multirow{3}{*}{3} & Baseline & 8.3 & \multirow{3}{*}{6} & Baseline & 89.2 \\
& AMS & 73.3 & & AMS & 90.3 \\
& Ours & \textbf{81.7} & & Ours & \textbf{90.9} \\
\midrule
\multirow{3}{*}{4} & Baseline & 27.2 & \multirow{3}{*}{7} & Baseline & \textbf{91.0}  \\
& AMS & 85.0 & & AMS & 90.7  \\
& Ours & \textbf{87.7} & & Ours & \textbf{91.0} \\
\midrule
\multirow{3}{*}{5} & Baseline & 80.5 & \multirow{3}{*}{$+\infty$} & \multirow{3}{*}{Baseline} & \multirow{3}{*}{91.6} \\
& AMS & 89.0 & & & \\
& Ours & \textbf{90.7} & & & \\
\bottomrule
\end{NiceTabular}
}
\end{small}
\end{center}
\vskip -2ex
\end{table}

%% file: tabs/tab_cifar_results.tex
\begin{table*}[t]
\caption{Accuracy with the 7-bit real chip (with the real chip curve from Figure~\ref{fig:adc_mapping} and noise level of 0.35) of bit-serial PIM system for different datasets and models. Note that PIM systems is hundreds times more efficient than software system.
}
\label{tab:cifar_results}
\vspace{-2.5ex}
\begin{center}
\begin{small}
\centering
\setlength{\tabcolsep}{5pt}
\setlength\extrarowheight{3pt}
\setlength{\aboverulesep}{0pt}
\setlength{\belowrulesep}{0pt}
\renewcommand{\arraystretch}{.9}
\scalebox{0.8}{
\begin{threeparttable}
\begin{tabular}{c|c|ccc|c|ccc|c|ccc}
\toprule
\multirow{1}{*}{Dataset} & \multirow{1}{*}{Model} & \multirow{1}{*}{Method} & \multirow{1}{*}{N} & \multirow{1}{*}{Acc.} & \multirow{1}{*}{Model} & \multirow{1}{*}{Method} & \multirow{1}{*}{N} & \multirow{1}{*}{Acc.} & \multirow{1}{*}{Model} & \multirow{1}{*}{Method} & \multirow{1}{*}{N} & \multirow{1}{*}{Acc.} \\
\hhline{-|-|---|-|---|-|---}
\multirow{10.3}{*}{CIFAR10} & \multirow{5.5}{*}{ResNet20} & Software & - & 91.6 & \multirow{5.5}{*}{ResNet44} & Software & - & 92.8 & \multirow{5.5}{*}{VGGNet11\tnote{$\dagger$}} & Software & - & 93.7 \\
\hhline{~|~|---|~|---|~|---}
& & \multirow{2}{*}{Baseline} & 72 & 13.9 & & \multirow{2}{*}{Baseline} & 72 & 10.5 & & \multirow{2}{*}{Baseline} & 72 & 10.0 \\
& & & 144 & 10.9 & & & 144 & 10.0 & & & 144 & 9.9 \\
\hhline{~|~|---|~|---|~|---}
& & \multirow{2}{*}{\textbf{Ours}} & 72 & \textbf{89.7} &  & \multirow{2}{*}{\textbf{Ours}} & 72 & \textbf{90.6} & & \multirow{2}{*}{\textbf{Ours}} & 72 & \textbf{94.2} \\
& & & 144 & \textbf{89.1}  & & & 144 & \textbf{90.7} & & & 144 & \textbf{94.0} \\
\hhline{~|-|---|-|---|-|---}
& \multirow{5.5}{*}{ResNet32} & Software & - & 92.5 & \multirow{5.5}{*}{ResNet56} & Software & - & 92.4 & \multicolumn{4}{l}{\multirow{5.5}{*}{}} \\
\hhline{~|~|---|~|---~~~~}
& & \multirow{2}{*}{Baseline} & 72 & 10.0 & & \multirow{2}{*}{Baseline} & 72 & 10.0 & \multicolumn{4}{l}{} \\
& & & 144 & 10.1 & & & 144 & 10.0 & \multicolumn{4}{l}{} \\
\hhline{~|~|---|~|---~~~~}
& & \multirow{2}{*}{\textbf{Ours}} & 72 & \textbf{90.6} & & \multirow{2}{*}{\textbf{Ours}} & 72 & \textbf{90.7} & \multicolumn{4}{l}{} \\
& & & 144 & \textbf{89.3} & & & 144 & \textbf{90.4} & \multicolumn{4}{l}{} \\
\hhline{=============}
\multirow{5.5}{*}{CIFAR100} 
& \multirow{5.5}{*}{ResNet20} & Software & - & 67.0 &
\multirow{5.5}{*}{ResNet56} & Software & - & 70.3 & \multirow{5.5}{*}{VGGNet11\tnote{$\dagger$}} & Software & - & NA \\
\hhline{~|~|---|~---~---}
& & \multirow{2}{*}{Baseline} & 72 & 1.8 & & \multirow{2}{*}{Baseline} & 72 & 1.0 & & \multirow{2}{*}{Baseline} & 72 & NA \\
& & & 144 & 1.3 & & & 144 & 1.1 & & & 144 & NA \\
\hhline{~|~|---|~---~---}
& & \multirow{2}{*}{\textbf{Ours}} & 72 & \textbf{62.6} & & \multirow{2}{*}{\textbf{Ours}} & 72 & \textbf{65.3} & & \multirow{2}{*}{\textbf{Ours}} & 72 & \textbf{NA} \\
& & & 144 & \textbf{61.8}  & & & 144 & \textbf{63.5} & & & 144 & \textbf{NA} \\
\bottomrule
\end{tabular}
\begin{tablenotes}\footnotesize
\item[$\dagger$] The architecture is the same as in~\citep{jia_programmable_2020}.
\item[*] Larger $N$ indicates higher efficiency but more information loss during quantization.
\end{tablenotes}
\end{threeparttable}
}
\end{small}
\end{center}
\vskip -3ex
\end{table*}

%% file: figs/fig_adjusted_bit_training.tex
\begin{figure}[htb]
\centering
\includegraphics[width=.6\linewidth]{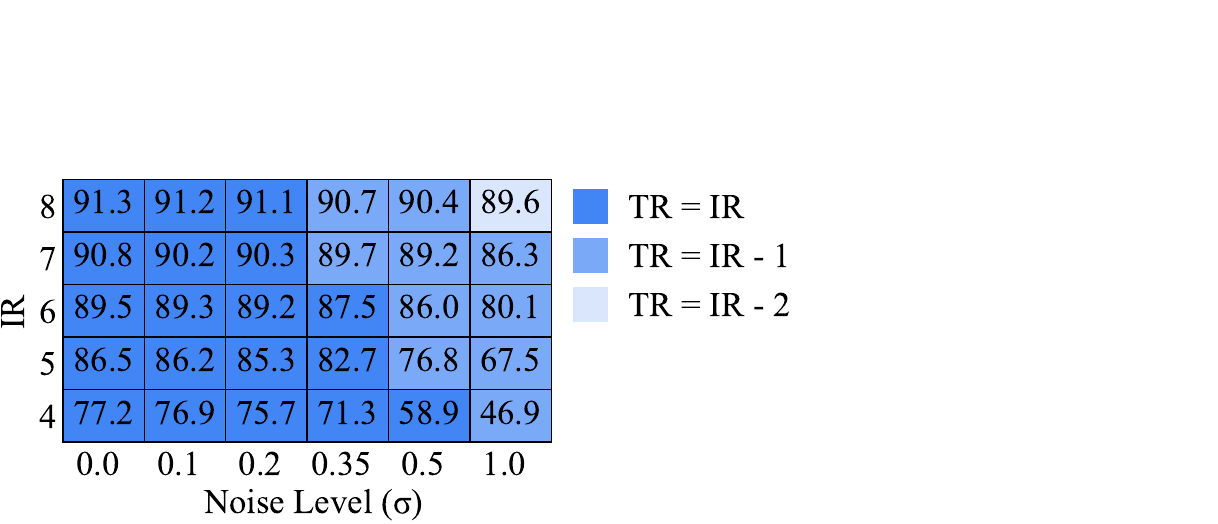}
\vskip -2ex
\caption{
The desirable training resolutions (TR) for different inference resolutions (IR) and noise levels, with bit-serial scheme (ResNet20 on CIFAR10). 
}
\label{fig:adjusted_bit_training}
\vskip -2ex
\end{figure}

%% file: figs/fig_noise.tex
\begin{figure*}[t]
\centering
\includegraphics[width=.9\linewidth]{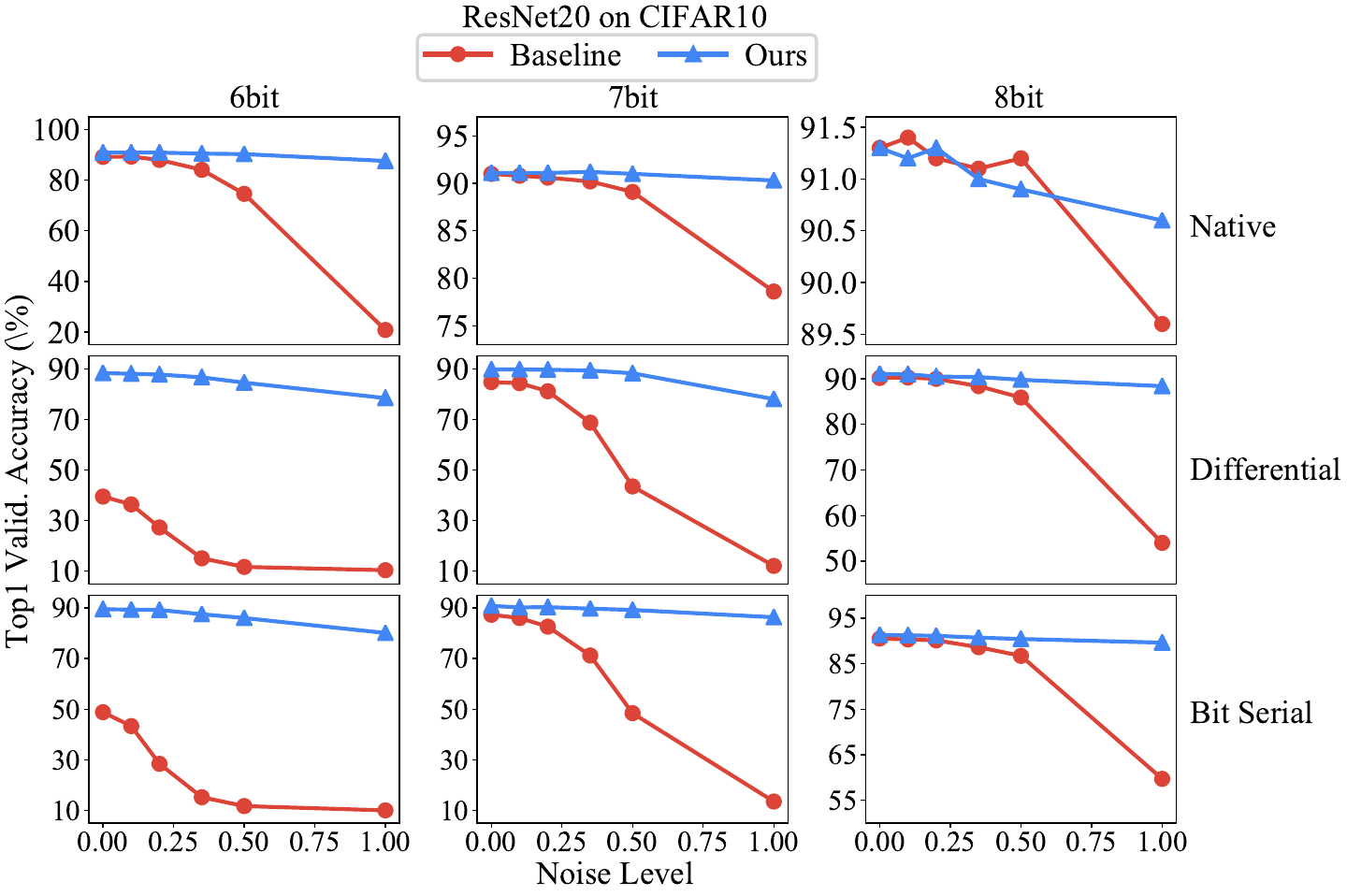}
\caption{
Performance of ResNet20 on CIFAR10 with ideal PIM of different schemes and PIM resolutions. 
Note that $N=9$ for native and $N=144$ for differential and bit serial schemes.
}
\label{fig:noise}
\vskip -2ex
\end{figure*}

%% file: tex_files/5_conclusion.tex
\section{Conclusion}
In this paper, we systematically study the problem of training a neural network for application on the processing in-memory (PIM) system, which is a promising candidate for next-generation hardware for deep learning, and we provide a method for the extra quantization step unique to PIM systems but ubiquitous to all different types of PIM implementations. Specifically, we formulate the problem and analyze the forward and backward propagation to enable PIM quantization-aware training. We study the training dynamics of our method, and propose rescaling techniques for both forward and backward propagations, to avoid gradient exploding/vanishing issues. We also study the discrepancy between training and inference, where more realistic non-ideal effects, such as imperfect linearity and stochastic thermal noise, are involved but difficult to incorporate during backward propagation. To this end, we propose to leverage BN calibration technique and invent adjusted precision training. Finally, we present experimental results to demonstrate potential relationship between training and inference bit-widths, together with noise level and effective number of bits for the PIM system for inference.




%% file: tex_files/appendix_into_main.tex
\clearpage


\setcounter{section}{0}
\setcounter{equation}{0}
\setcounter{figure}{0}
\setcounter{table}{0}

\renewcommand{\theequation}{A\arabic{equation}}
\renewcommand{\thesection}{A\arabic{section}}
\renewcommand{\thefigure}{A\arabic{figure}}
\renewcommand{\thetable}{A\arabic{table}}



\input{tex_files/A1_proof_two_column}

\input{tex_files/A2_setting}
\input{tex_files/A3_scale_enlarging}
\input{tex_files/A4_running_stats_nonideality}
\input{tex_files/A5_scaling_factor}
\input{tex_files/A6_ablation}

\input{tex_files/A7_bn_calib}

%% file: tex_files/A1_proof_two_column.tex
\section{Proof of Theorems}
\label{sec:proof}

Here we present detailed proofs for Theorem 1 and 2. We first formulate the quantization procedure of PIM systems with several popular schemes, then derive the results of Theorem 1. After that, we analyze the training dynamics of a generic neural networks to prove Theorem 2.

\subsection{PIM Quantization-Aware Training}
To prove Theorem 1, we first present the quantization procedure of PIM systems with native, differential and bit serial schemes. The output of a linear layer is given by
\begin{equation}
    \widetilde{y}=\sum_{i=1}^N\widetilde{Q}_i\widetilde{q}_i
\end{equation}
where $\widetilde{Q}_i\in[-1,1]$ is quantized weight of $b_w$ bits and $\widetilde{q}_i\in[0,1]$ is quantized input of $b_a$ bits, respectively.

For PIM systems, due to the limited resolution of digital-to-analog converter (which is $m$ bits) for the inputs, the inputs are first decomposed into sub-arrays of $m$ bits. In other words, we have
\begin{subequations}
\allowdisplaybreaks
\begin{align}
    \widetilde{q}_i&=\sum_{k=0}^{b_a/m-1}\widetilde{q}^{(m)}_{i,k}\Delta^k\\
    \widetilde{q}^{(m)}_{i,k}&=\frac{1}{2^{b_a}-1}q^{(m)}_{i,k}\\
    \Delta&=2^m
\end{align}
\end{subequations}
with $q^{(m)}_{i,k}\in\{0,1,\dots,\Delta-1\}$.

\paragraph{Native Scheme} For PIM system with native scheme, the output is given by
\begin{subequations}
\allowdisplaybreaks
\begin{align}
    \widetilde{y}_{\IMC}&=\bm{\mathsf{Q}}\bigg(\sum_{i=1}^N\widetilde{Q}_i\widetilde{q}_i;b_{\IMC}\bigg)\\
    &=\sum_{k=0}^{b_a/m-1}\Delta^k\cdot\frac{N(\Delta-1)}{(2^{b_{\IMC}}-1)(2^{b_a}-1)}\nonumber\\
    &\cdot\round\Big(\frac{(2^{b_{\IMC}}-1)(2^{b_a}-1)}{N(\Delta-1)}\sum_{i=1}^N\widetilde{Q}_i\cdot \widetilde{q}^{(m)}_{i,k}\Big)\label{eq:native}
\end{align}
\end{subequations}
With the GSTE assumption, we can derive its differential as
\begin{subequations}
\allowdisplaybreaks
\begin{align}
    \mathrm{d}\widetilde{y}_{\IMC}&=\sum_{k=0}^{b_a/m-1}\Delta^k\cdot\frac{N(\Delta-1)}{(2^{b_{\IMC}}-1)(2^{b_a}-1)}\nonumber\\
    &\cdot\mathrm{d}\Big[\round\Big(\frac{(2^{b_{\IMC}}-1)(2^{b_a}-1)}{N(\Delta-1)}\sum_{i=1}^N\widetilde{Q}_i\cdot \widetilde{q}^{(m)}_{i,k}\Big)\Big]\\
    &=\sum_{k=0}^{b_a/m-1}\Delta^k\cdot\frac{N(\Delta-1)}{(2^{b_{\IMC}}-1)(2^{b_a}-1)}\cdot\xi\nonumber\\
    &\quad\cdot\mathrm{d}\Big(\frac{(2^{b_{\IMC}}-1)(2^{b_a}-1)}{N(\Delta-1)}\sum_{i=1}^N\widetilde{Q}_i\cdot \widetilde{q}^{(m)}_{i,k}\Big)\\
    &=\xi\cdot\mathrm{d}\Big(\sum_{i=1}^N\widetilde{Q}_i\cdot \sum_{k=0}^{b_a/m-1}\widetilde{q}^{(m)}_{i,k}\Delta^k\Big)\\
    &=\xi\cdot\mathrm{d}\bigg(\sum_{i=1}^N\widetilde{Q}_i\widetilde{q}_i\bigg)\label{eq:native_diff}
\end{align}
\end{subequations}

\paragraph{Differential Scheme} For PIM system with differential scheme, the weight is first decomposed into positive and negative parts, as
\begin{equation}
    \widetilde{Q}_i=\widetilde{Q}_i^+ +\widetilde{Q}_i^-
\end{equation}
where all elements in $\widetilde{Q}_i^+$ are positive and those in $\widetilde{Q}_i^-$ are negative. Its differential is given by
\begin{equation}
    \mathrm{d}\widetilde{Q}_i=\mathrm{d}\widetilde{Q}_i^+ +\mathrm{d}\widetilde{Q}_i^-
\end{equation}
The output is the combination of these two parts as
\begin{subequations}
\allowdisplaybreaks
\begin{align}
    \widetilde{y}_{\IMC}&=\bm{\mathsf{Q}}\bigg(\sum_{i=1}^N\widetilde{Q}_i\widetilde{q}_i;b_{\IMC}\bigg)\\
    &=\sum_{k=0}^{b_a/m-1}\Delta^k\cdot\frac{N(\Delta-1)}{(2^{b_{\IMC}}-1)(2^{b_a}-1)}\nonumber\\
    &\quad\cdot\Big[\round\Big(\frac{(2^{b_{\IMC}}-1)(2^{b_a}-1)}{N(\Delta-1)}\sum_{i=1}^N\widetilde{Q}_i^+\widetilde{q}^{(m)}_{i,k}\Big)\nonumber\\
    &\quad-\round\Big(\frac{(2^{b_{\IMC}}-1)(2^{b_a}-1)}{N(\Delta-1)}\sum_{i=1}^N(-\widetilde{Q}_i^-)\widetilde{q}^{(m)}_{i,k}\Big)\Big]\label{eq:differential}
\end{align}
\end{subequations}
Taking differential on both sides gives
\begin{subequations}
\allowdisplaybreaks
\begin{align}
    \mathrm{d}\widetilde{y}_{\IMC}&=\mathrm{d}\sum_{k=0}^{b_a/m-1}\Delta^k\cdot\frac{N(\Delta-1)}{(2^{b_{\IMC}}-1)(2^{b_a}-1)}\nonumber\\
    &\cdot\Big[\round\Big(\frac{(2^{b_{\IMC}}-1)(2^{b_a}-1)}{N(\Delta-1)}\sum_{i=1}^N\widetilde{Q}_i^+\widetilde{q}^{(m)}_{i,k}\Big)\nonumber\\
    &-\round\Big(\frac{(2^{b_{\IMC}}-1)(2^{b_a}-1)}{N(\Delta-1)}\sum_{i=1}^N(-\widetilde{Q}_i^-)\widetilde{q}^{(m)}_{i,k}\Big)\Big]\\
    &=\sum_{k=0}^{b_a/m-1}\Delta^k\cdot\frac{N(\Delta-1)}{(2^{b_{\IMC}}-1)(2^{b_a}-1)}\nonumber\\
    &\cdot\Big\{\mathrm{d}\Big[\round\Big(\frac{(2^{b_{\IMC}}-1)(2^{b_a}-1)}{N(\Delta-1)}\sum_{i=1}^N\widetilde{Q}_i^+\widetilde{q}^{(m)}_{i,k}\Big)\Big]\nonumber\\
    &-\mathrm{d}\Big[\round\Big(\frac{(2^{b_{\IMC}}-1)(2^{b_a}-1)}{N(\Delta-1)}\sum_{i=1}^N(-\widetilde{Q}_i^-)\widetilde{q}^{(m)}_{i,k}\Big)\Big]\Big\}\\
    &=\sum_{k=0}^{b_a/m-1}\Delta^k\cdot\frac{N(\Delta-1)}{(2^{b_{\IMC}}-1)(2^{b_a}-1)}\nonumber\\
    &\cdot\Big[\xi\cdot\mathrm{d}\Big(\frac{(2^{b_{\IMC}}-1)(2^{b_a}-1)}{N(\Delta-1)}\sum_{i=1}^N\widetilde{Q}_i^+\widetilde{q}^{(m)}_{i,k}\Big)\nonumber\\
    &-\xi\cdot\mathrm{d}\Big(\frac{(2^{b_{\IMC}}-1)(2^{b_a}-1)}{N(\Delta-1)}\sum_{i=1}^N(-\widetilde{Q}_i^-)\widetilde{q}^{(m)}_{i,k}\Big)\Big]\\
    &=\xi\cdot\mathrm{d}\Big(\sum_{i=1}^N\widetilde{Q}_i^+\sum_{k=0}^{b_a/m-1}\widetilde{q}^{(m)}_{i,k}\Delta^k\Big)\nonumber\\
    &+\xi\cdot\mathrm{d}\Big(\sum_{i=1}^N\widetilde{Q}_i^-\sum_{k=0}^{b_a/m-1}\widetilde{q}^{(m)}_{i,k}\Delta^k\Big)\\
    &=\xi\cdot\mathrm{d}\Big(\sum_{i=1}^N\widetilde{Q}_i^+\widetilde{q}_i\Big)+\xi\cdot\mathrm{d}\Big(\sum_{i=1}^N\widetilde{Q}_i^-\widetilde{q}_i\Big)\\
    &=\xi\cdot\mathrm{d}\Big(\sum_{i=1}^N(\widetilde{Q}_i^++\widetilde{Q}_i^-)\widetilde{q}_i\Big)\\
    &=\xi\cdot\mathrm{d}\bigg(\sum_{i=1}^N\widetilde{Q}_i\widetilde{q}_i\bigg)\label{eq:differential_diff}
\end{align}
\end{subequations}

\paragraph{Bit Serial Scheme} For PIM system with bit serial scheme, the weight is first decomposed into bits as
\begin{equation}
    \widetilde{Q}_i=\sum_{k=0}^{b_w-1}(-1)^{\delta_{k,b_w-1}}\widetilde{Q}_{i,k}2^k
\end{equation}
where
\begin{equation}
    \widetilde{Q}_{i,k}=\frac{1}{2^{b_w-1}-1}Q_{i,k}
\end{equation}
and $Q_{i,k}\in\{0,1\}$.
The output is obtained for each bits separately and then summed over as
\begin{subequations}
\allowdisplaybreaks
\begin{align}
    \widetilde{y}_{\IMC}&=\bm{\mathsf{Q}}\bigg(\sum_{i=1}^N\widetilde{Q}_i\widetilde{q}_i;b_{\IMC}\bigg)\\
    &=\sum_{k=0}^{b_w-1}\sum_{l=0}^{b_a/m-1}(-1)^{\delta_{k,b_w-1}}2^k\Delta^l\nonumber\\
    &\cdot\frac{N(\Delta-1)}{(2^{b_{\IMC}}-1)(2^{b_w-1}-1)(2^{b_a}-1)}\nonumber\\
    &\cdot\round\Big(\frac{(2^{b_{\IMC}}-1)(2^{b_w-1}-1)(2^{b_a}-1)}{N(\Delta-1)}\nonumber\\
    &\cdot\sum_{i=1}^N\widetilde{Q}_{i,k}\widetilde{q}^{(m)}_{i,l}\Big)\label{eq:bit_serial}
\end{align}
\end{subequations}
The differential can thus be determined as
\begin{subequations}
\allowdisplaybreaks
\begin{align}
    \mathrm{d}\widetilde{y}_{\IMC}&=\mathrm{d}\sum_{k=0}^{b_w-1}\sum_{l=0}^{b_a/m-1}(-1)^{\delta_{k,b_w-1}}2^k\Delta^l\nonumber\\
    &\quad\cdot\frac{N(\Delta-1)}{(2^{b_{\IMC}}-1)(2^{b_w-1}-1)(2^{b_a}-1)}\nonumber\\
    &\quad\cdot\round\Big(\frac{(2^{b_{\IMC}}-1)(2^{b_w-1}-1)(2^{b_a}-1)}{N(\Delta-1)}\nonumber\\
    &\quad\sum_{i=1}^N\widetilde{Q}_{i,k}\widetilde{q}^{(m)}_{i,l}\Big)\\
    &=\sum_{k=0}^{b_w-1}\sum_{l=0}^{b_a/m-1}(-1)^{\delta_{k,b_w-1}}2^k\Delta^l\nonumber\\
    &\quad\cdot\frac{N(\Delta-1)}{(2^{b_{\IMC}}-1)(2^{b_w-1}-1)(2^{b_a}-1)}\nonumber\\
    &\quad\cdot\mathrm{d}\Big[\round\Big(\frac{(2^{b_{\IMC}}-1)(2^{b_w-1}-1)(2^{b_a}-1)}{N(\Delta-1)}\nonumber\\
    &\quad\sum_{i=1}^N\widetilde{Q}_{i,k}\widetilde{q}^{(m)}_{i,l}\Big)\Big]\\
    &=\sum_{k=0}^{b_w-1}\sum_{l=0}^{b_a/m-1}(-1)^{\delta_{k,b_w-1}}2^k\Delta^l\nonumber\\
    &\quad\cdot\frac{N(\Delta-1)}{(2^{b_{\IMC}}-1)(2^{b_w-1}-1)(2^{b_a}-1)}\nonumber\\
    &\quad\cdot\xi\cdot\mathrm{d}\Big(\frac{(2^{b_{\IMC}}-1)(2^{b_w-1}-1)(2^{b_a}-1)}{N(\Delta-1)}\nonumber\\
    &\quad\cdot\sum_{i=1}^N\widetilde{Q}_{i,k}\widetilde{q}^{(m)}_{i,l}\Big)\\
    &=\xi\cdot\mathrm{d}\Big(\sum_{i=1}^N\sum_{k=0}^{b_w-1}(-1)^{\delta_{k,b_w-1}}\widetilde{Q}_{i,k}2^k\nonumber\\
    &\quad\cdot\sum_{l=0}^{b_a/m-1}\widetilde{q}^{(m)}_{i,l}\Delta^l\Big)\\
    &=\xi\cdot\mathrm{d}\bigg(\sum_{i=1}^N\widetilde{Q}_i\widetilde{q}_i\bigg)\label{eq:bit_serial_diff}
\end{align}
\end{subequations}

With~\eqref{eq:native_diff},~\eqref{eq:differential_diff} and~\eqref{eq:bit_serial_diff}, we prove Theorem 1.

\subsection{Training Dynamics}

Here we analyze the training dynamics to prove Theorem 2. Our analysis is similar to that presented in~\citep{quantefficienttraining}, which is zeroth order approximation and is based on mean field theory, where different quantities are assumed to be independent (although some of them have some dependence, especially the gradients, as described following Axiom 3.2 in~\citep{yang2017mean}).

We want to analyze the training dynamics of a generic neural network with repeated blocks composed of linear layer, some nonlinear effect, forward scaling, batch normalization, and output activation function, as
\begin{subequations}
\allowdisplaybreaks
\begin{align}
    x^{(l+1)}_i &= \varphi(y^{(l)}_i)\\
    y^{(l)}_i &= \gamma^{(l)}_i\frac{z^{(l)}_i-\mu^{(l)}_i}{\sigma^{(l)}_i} + \beta^{(l)}_i\\
    z^{(l)}_i &=\eta^{(l)}\widetilde{z}^{(l)}_i\\
    \widetilde{z}^{(l)}_i &= f(W^{(l)}_{ij},x^{(l)}_j)\\
    &\sim\rho^{(l)}\sum_{j=1}^{n^{(l)}}W^{(l)}_{ij}x^{(l)}_j\\
    \mathrm{d}\widetilde{z}^{(l)}_i &=\xi^{(l)}\cdot\mathrm{d}\left(\sum_{j=1}^{n^{(l)}}W^{(l)}_{ij}x^{(l)}_j\right)
\end{align}
\end{subequations}
where $\rho^{(l)}$ denotes modification on the output variance by the nonlinear effect and $\xi^{(l)}$ is the scaling factor introduced during backward propagation. From this we can derive the following statistics

\begin{subequations}
\allowdisplaybreaks
\begin{align}
    \VAR{y^{(l)}_i}&\approx(\gamma^{(l)}_i)^2\\
    \E{(x^{(l)}_i)^2}&\approx\E{(\varphi'(y^{(l)}_i))^2}\VAR{y^{(l)}_i}\label{eq:mse}\\
    &\approx\E{(\varphi'(y^{(l)}_i))^2}(\gamma^{(l)}_i)^2\\
    (\sigma^{(l)}_i)^2&=\VAR{z^{(l)}_i}\\
    &=(\eta^{(l)})^2(\rho^{(l)})^2n^{(l)}\VAR{W^{(l)}_{ij}}\E{(x^{(l)}_j)^2}\\
    &\approx (\eta^{(l)})^2(\rho^{(l)})^2n^{(l)}\VAR{W^{(l)}_{ij}}\nonumber\\
    &\quad\cdot\E{(\varphi'(y^{(l)}_j))^2}(\gamma^{(l)}_j)^2\label{eq:sigma}
\end{align}
\end{subequations}
where~\eqref{eq:mse} is valid if the activation $\phi$ is quasi-linear.

We first estimate the gradient of the batch normalization layer as following
\begin{subequations}
\allowdisplaybreaks
\begin{align}
    \frac{\partial\mu^{(l)}_i}{\partial z^{(l)}_i}&=\frac{1}{m_B}\\
    \frac{\partial\sigma^{(l)}_i}{\partial z^{(l)}_i}&=\frac{1}{m_B}\frac{z^{(l)}_i-\mu^{(l)}_i}{\sigma^{(l)}_i}\\
    \frac{\partial y^{(l)}_i}{\partial z^{(l)}_i}&=\gamma^{(l)}_i\left[\frac{1}{\sigma^{(l)}_i}-\frac{1}{\sigma^{(l)}_i}\frac{\partial\mu^{(l)}_i}{\partial z^{(l)}_i}\right.\nonumber\\
    &\left.\quad+(z^{(l)}_i-\mu^{(l)}_i)\frac{-1}{(\sigma^{(l)}_i)^2}\frac{\partial\sigma^{(l)}_i}{\partial z^{(l)}_i}\right]\\
    &=\gamma^{(l)}_i\left[\frac{1}{\sigma^{(l)}_i}-\frac{1}{\sigma^{(l)}_i}\frac{1}{m_B}\right.\nonumber\\
    &\left.\quad-\frac{1}{\sigma^{(l)}_i}\frac{1}{m_B}\left(\frac{z^{(l)}_i-\mu^{(l)}_i}{\sigma^{(l)}_i}\right)^2\right]\\
    &\approx\frac{\gamma^{(l)}_i}{\sigma^{(l)}_i}\quad(m_B>>1)
\end{align}
\end{subequations}
where we have assumed that the batch size is large enough, which is typically satisfied in practice.

The gradients of loss $\mathcal{L}$ with respect to the input can be easily calculated, which is
\begin{equation}
    \partial_{x^{(l)}_j}\mathcal{L}=\xi^{(l)}\sum_{i=1}^{n^{(l+1)}}W^{(l)}_{ij}\eta^{(l)}\frac{\gamma^{(l)}_i}{\sigma^{(l)}_i}\varphi'(y^{(l)}_i)\partial_{x^{(l+1)}_i}\mathcal{L}
\end{equation}
from which we can derive the variance of the gradient, based on mean field assumption, as
\begin{subequations}
\begin{align}
    \VAR{\partial_{x^{(l)}_j}\mathcal{L}}&=n^{(l+1)}(\xi^{(l)})^2(\eta^{(l)})^2\left(\frac{\gamma^{(l)}_i}{\sigma^{(l)}_i}\right)^2\nonumber\\
    &\quad\cdot\VAR{W^{(l)}_{ij}}\E{(\varphi'(y^{(l)}_i))^2}\VAR{\partial_{x^{(l+1)}_i}\mathcal{L}}
\end{align}
\end{subequations}
Substituting~\eqref{eq:sigma}, we have
\begin{subequations}
\allowdisplaybreaks
\begin{align}
    &\VAR{\partial_{x^{(l)}_j}\mathcal{L}}\nonumber\\
    =&n^{(l+1)}(\xi^{(l)})^2\cancel{(\eta^{(l)})^2}\nonumber\\
    &\cdot\frac{\cancel{(\gamma^{(l)}_i)^2}}{n^{(l)}\cancel{(\eta^{(l)})^2}(\rho^{(l)})^2\cancel{\VAR{W^{(l)}_{ij}}}\cancel{\E{(\varphi'(y^{(l)}_j))^2}}\cancel{(\gamma^{(l)}_j)^2}}\nonumber\\
    &\cdot\cancel{\VAR{W^{(l)}_{ij}}}\cdot \cancel{\E{(\varphi'(y^{(l)}_i))^2}}\VAR{\partial_{x^{(l+1)}_i}\mathcal{L}}\nonumber\\
    =&\left(\frac{\xi^{(l)}}{\rho^{(l)}}\right)^2\cdot\frac{n^{(l+1)}}{n^{(l)}}\cdot\VAR{\partial_{x^{(l+1)}_i}\mathcal{L}}
\end{align}
\end{subequations}


Ignoring spatial dependence of all statistics, we have
\begin{equation}
\boxed{
    \frac{\VAR{\partial_{x^{(l)}}\mathcal{L}}}{\VAR{\partial_{x^{(l+1)}}\mathcal{L}}}\approx\left(\frac{\xi^{(l)}}{\rho^{(l)}}\right)^2\cdot\frac{n^{(l+1)}}{n^{(l)}}
}
\end{equation}

%% file: tex_files/A2_setting.tex
\section{Experiment Settings}
\label{sec:detail_settings}



\input{figs/fig_adc_table}

\subsection{General Experiment Settings}
\label{sec:exp_settings}
Our method is evaluated using
several ResNet models on CIFAR. Weights and inputs are quantized to $4$-bit, and 
$b_{\IMC}$ varies from $3$ to $10$. The first convolution layer and the final fully-connection layer for classification are implemented on digital system, namely $b_{\IMC}=+\infty$ for these two layers. 
{To accurately evaluate the inference accuracy on actual hardware with variations, non-linearity, and noise, we evaluate the proposed method using physical models of a state-of-the-art SRAM PIM chip prototype. Each PIM SRAM macro in the chip computes 32 analog MACs ($N<=144$, $b_{\IMC}$=7) in parallel. The measured 32 transfer functions, shown in 
Fig.~\ref{fig:adc_mapping}, 
capture all the nonlinearity and mismatch of the physical chip. Random noise in computation, which follows Gaussian distribution and is solely characterized by root-mean-square (RMS) error~\citep{gray2009analysis,razavi2016design,sansen2007analog,pelgrom2013analog,maloberti2007data}, is measured to be 0.35 LSB. Due to the small size of the prototype chip, running through all images in test dataset is infeasible in time, so we build a hardware calibrated physical model to quickly, accurately and flexibly simulate the inference accuracy of real hardware, which is a widely adopted common practice in custom hardware research because of the inhibiting costs of building a full-scale chip for large DNNs~\citep{yue_143_2020,su_152_2020,si_155_2020,jia_programmable_2020,jiang2019c3sram}. We have experimentally confirmed the identical MAC and inference results of the model and a real physical chip. It is also worth noting that the non-idealities presented in this SRAM PIM chip is representative of that of various types of PIM hardware.
}

To verify the effectiveness of our method, we experiment on ResNet20, ResNet32, ResNet44 and ResNet56 on CIFAR10 and ResNet20 on CIFAR100. Following previous practice~\citep{sat}, weights and inputs of convolution and fully-connected layers are quantized to $4$-bit ($b_w=b_a=4$), including the first and last layers, except that inputs to the first layer are kept at $8$ bit, and we do not apply normalization on these images. Batch normalization layers and bias in the final fully-connected layers are full-precision. The quantization resolution for PIM system ($b_{\IMC}$) varies from $3$ to $10$. The first convolution layer and the final fully-connection layer for classification are implemented on digital system, namely $b_{\IMC}=+\infty$ for these two layers. For CIFAR10 and CIFAR100, the $1\times1$ convolution layers for residual connection require much less computations and thus are also implemented on digital system. For differential and bit serial scheme, inputs are first split along the channel dimension into sub-tensors, each with a unit channel of $16$, corresponding to $N=144$ for $3\times3$ convolution, and processed separately before summing the final PIM outputs. For native scheme, we instead use a unit channel of $1$ and thus $N=9$ to match the experiment setting in~\cite{ams} which use $N=8$. For real-curve results, since we totally have $32$ curves (ADC components), each for $8$ outputs with $b_w=4$ bits, the output channels are split with unit output channel of $8$.

The input image is randomly cropped to $32\times32$ and randomly flipped horizontally during training and directly applied without augmentation for inference. All models are trained from scratch with $200$ epochs and multi-step learning rate scheduler, where the initial learning rate is $0.1$ and reduced by $10$ times at the $100$-th and $150$-th epochs. We use SGD optimizer with Nesterov momentum of weight 0.9 and weight decay is set to $0.0001$. Batch size is $128$ for all experiments. We apply constant rescaling on all layers, including the convolution layers, in contrast to only the last fully-connected layers suggested in~\citep{quantefficienttraining}, despite batch normalization is applied in the model. All experiments are finished on one GeForce GTX 1080 GPU with $12$GB memory.

Weights are quantized with a modified DoReFa scheme, without mapping between intervals of $[-1,1]$ and $[0,1]$. Specifically, the quantized weights are given as
\begin{subequations}
\begin{align}
    Q_i&=\frac{1}{2^{b_w-1}-1}s\cdot\round\Big((2^{b_w-1}-1)\nonumber\\
    &\qquad\cdot\frac{\tanh(W_i)}{\max\limits_{k}|\tanh(W_k)|}\Big)\label{eq:nudge_scheme}\\
    s&=\frac{1}{\sqrt{n_{\mathrm{out}}\mathbb{VAR}[Q_{i}]}}
\end{align}
\end{subequations}
where $n_{\mathrm{out}}$ denotes the number of output neurons of the linear layer. For native scheme, since the output can also be negative, we also adopt such quantization function.

\subsection{Error Analysis Experiment Settings}



\paragraph{Computing Error Analysis (Fig.~\ref{fig:enob_nonideality})} For this example, we first obtain the MAC results via uniform random sampling on the output space, and apply PIM quantization together with noise injection. By comparing the ideal output with the noisy quantized output for different noise levels, we can obtain the errors, from which we estimate the standard deviation of them, for each value of noise levels. These standard deviations are then normalized by that for the noiseless quantization.



%% file: figs/fig_adc_table.tex
\begin{figure}[t]
\vskip -2ex
\centering
\includegraphics[width=.7\linewidth]{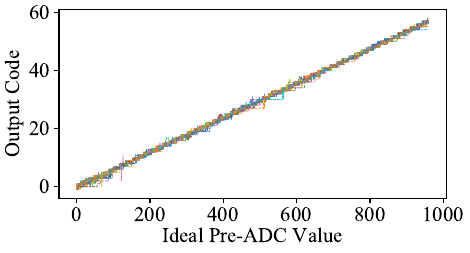}
\caption{
Imperfect MAC outputs from a PIM chip.
}
\label{fig:adc_mapping}
\vskip -2ex
\end{figure}

%% file: tex_files/A3_scale_enlarging.tex
\section{Scale-Enlarging Effect of PIM Quantization}
\label{sec:scale_enlarging}

\input{figs/fig_bit_adc_std_ratio}

Here we show the scale-enlarging effect of PIM quantization. Specifically, we study an idealized noiseless system to examine the effect of $\bm{\mathsf{Q}}$ on the standard deviation of outputs. As an example, we experiment on a toy example of convolution 
with bit serial scheme, and calculate standard deviation ratio between the outputs with and without PIM quantization. 
For this purpose, we set the number of input channels to $16$ and that of output channels to $32$. The kernel size is given by $3\times3$, and both inputs and weights are quantized to $4$ bit. We experiment on a random batch of $100$ data, each distributed uniformly on $[0, 1]$ before quantized. Weights are randomly sampled with normal distribution under Kaiming intialization condition~\citep{he2015delving}, and quantized with the previously mentioned modified DoReFa scheme~\citep{dorefa}, given by~\eqref{eq:nudge_scheme}. We experiment on an ideal bit-serial PIM system, and calculate standard deviation ratio between the output of PIM system and that from conventional quantization with digital accelerator.
We plot this ratio against PIM bit-width ($b_{\IMC}$), and obtain such curves for different numbers of input channels. As illustrated in Figure~\ref{fig:std_ratio}, we find that the difference between the two scenarios is not significant for high precision, which is as expected. However, for mediate precision such as $5\sim7$ bits, they start to become different, and for ultra-low bit-widths, such as $3\sim4$ bits, the discrepancy can be as large as $2\sim4$. 


%% file: figs/fig_bit_adc_std_ratio.tex
\begin{figure}[t]
\centering
\includegraphics[width=.7\linewidth]{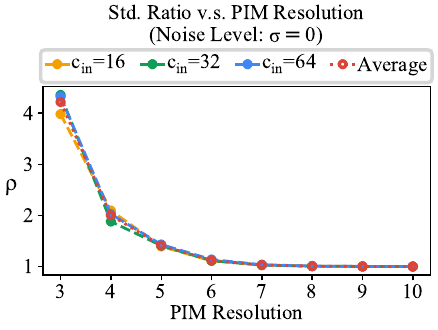}
\caption{
Impact of PIM quantization on the output scale measured by standard deviation. $\rho$ is defined as in~\eqref{eq:scale_nonlinear}.
}
\label{fig:std_ratio}
\vskip -3ex
\end{figure}

%% file: tex_files/A4_running_stats_nonideality.tex
\section{Impact of Non-idealities on BN Statistics}
\input{figs/fig_running_stats_nonideality}
\label{sec:bn_stats_nonideality}

In this section, we study the impact of non-ideal effects on BN statistics. We experiment with a toy example of one layer convolution implemented on ideal or real PIM systems, and calculate the running statistics of output for different noise levels.
For this purpose, we use the same toy experiment setting as in~\ref{sec:scale_enlarging}, and set the unit output channel for real-curve inference to $8$. 
The results are illustrated in Figure~\ref{fig:running_stats_nonideality}. We find that 
output statistics can change by as much as $30\%$, which might have significant impact on the model's final output, especially if its behavior is sensitive to these values.

%% file: figs/fig_running_stats_nonideality.tex
\begin{figure}[t]
\centering
\includegraphics[width=.7\linewidth]{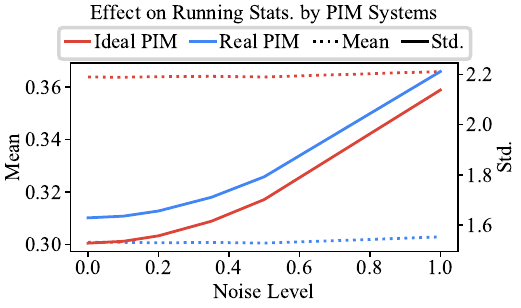}
\caption{
Impact of nonlinearity and noise non-idealities on the running statistics. Note this is one sampling results.
}
\label{fig:running_stats_nonideality}
\end{figure}

%% file: tex_files/A5_scaling_factor.tex
\section{Scaling Factors for Forward Rescaling}

Here we list the rescaling factors for forward propagation, as shown in Table~\ref{tab:scale_factor}. We find that it depends on PIM resolution $b_{\IMC}$ and PIM decomposition scheme. Moreover, it can even be different for different software package versions. As mentioned in the text, the underlying reason is still unclear to us.

\input{tabs/tab_scale_factor}

%% file: tabs/tab_scale_factor.tex
\begin{table}[t]
\caption{Scaling factor for forward rescaling for different PIM resolution $b_{\IMC}$ and different PIM decomposition schemes.}
\label{tab:scale_factor}
\begin{center}
\begin{small}
\setlength{\tabcolsep}{3.5pt}
\renewcommand{\arraystretch}{1.2}
\begin{tabular}{cccc}
\toprule
$b_{\IMC}$ & Native & Differential & Bit Serial \\
\midrule
3 & 100 & 1000 & 100 \\
4 & 20  & 1000 & 30 \\
5 & 1  & 1000 & 30 \\
6 & 1  & 1000 & 30 \\
7 & 1  & 1000 & 1.03 \\
\bottomrule
\end{tabular}
\end{small}
\end{center}
\end{table}

%% file: tex_files/A6_ablation.tex
\section{Ablation Study}
Here we provide more in-depth ablation study for our methods, including of PIM quantization-aware training, the rescaling techniques, and batch normalization calibration.

\subsection{PIM-QAT and Rescaling}
We first study the effectiveness of PIM-QAT together with the rescaling techniques. To this end, we compare the baseline with that trained with PIM-QAT, without using BN calibration or adjusted bit training. The two rescaling techniques for forward and backward propagation are applied, and we use noiseless ideal PIM system, without using any real chip curve. Table~\ref{tab:framework} compares the results of bit serial scheme for several different $b_{\IMC}$'s, which are also plotted in Figure~\ref{fig:adc_quant_rescaling}. It can be seen that for low $b_{\IMC}$, our method is significantly better than the baseline. 
Specifically, for $b_{\IMC}<9$, our method gives better results, and for ultra-low bit-width, such as $3$ bit, where 
baseline models are not different from random guess, our method can still get a reasonable top-1 accuracy of $61.8\%$. We also find that for sufficient high $b_{\IMC}$ (larger than $8$), baseline can be better.
This is also reasonable as the noiseless PIM with such high precision will almost introduce no precision loss.

\input{tabs/tab_framework}
\input{figs/fig_adc_quant_and_rescaling}

\subsection{Rescaling}
We then study the rescaling techniques we propose for both forward and backward propagations. As listed in Table~\ref{tab:rescaling} and shown in Fig.~\ref{fig:rebuttal_training_curves}, for bit serial scheme, if the $b_{\IMC}$ is lower than $6$, training without forward or backward propagation will all make the training unstable.
These experiments demonstrate that both rescaling techniques we propose are necessary and beneficial for stablized training dynamics of the neural network.
Experiments on native and differential schemes give similar results.

\input{tabs/tab_rescaling}
\input{figs/fig_rebuttal_training_curve}

\subsection{BN Calibration}

\input{figs/fig_bn_calib}

Besides training techniques discussed above, the discrepancy between training with idealized quantization and inference with real-case non-idealities, including non-linearity and noise, are dealt with BN calibration. To verify its effectiveness, we compare the results using the BN calibration or not for both baseline and our method, and illustrate the results for $7$ bit ideal and real PIM in Figure~\ref{fig:bn_calib}. 
We find that BN calibration significantly improves the results for all cases, especially for those with real PIM system. More interestingly, it also improves the baseline results, yet the performance is still unsatisfactory and significantly worse than ours. These experiments demonstrate that the change of BN running statistics caused by nonlinearity and noise effects of real PIM systems has strong impact on the predictive capability of the neural network, and this can be alleviated to a large extent with a simple yet effective software solution, without extra training efforts.

%% file: tabs/tab_framework.tex
\begin{table}[htb!]
\caption{Accuracy of ResNet20 on CIFAR10 with idealized bit-serial PIM-quantization, where noise or real chip curve are not involved. For our results, we use rescaling techniques for both forward and backward propagation, as described in the text.}
\label{tab:framework}
\begin{center}
\begin{small}
\begin{NiceTabular}{ccc|ccc}
\toprule
$b_{\IMC}$ & Method & Acc. & $b_{\IMC}$ & Method & Acc. \\
\midrule
\multirow{2}{*}{3} & Baseline & 10.0 & \multirow{2}{*}{7} & Baseline & 85.8 \\
& \textbf{Ours} & \textbf{61.8} & & \textbf{Ours} & \textbf{90.8} \\
\midrule
\multirow{2}{*}{4} & Baseline & 10.2 & \multirow{2}{*}{8} & Baseline & 90.3 \\
& \textbf{Ours} & \textbf{77.2} & & \textbf{Ours} & \textbf{90.8} \\
\midrule
\multirow{2}{*}{5} & Baseline & 11.0 & \multirow{2}{*}{9} & Baseline & \textbf{91.2} \\
& \textbf{Ours} & \textbf{86.5} & & Ours & 90.8 \\
\midrule
\multirow{2}{*}{6} & Baseline & 41.1 & \multirow{2}{*}{10} & \textbf{Baseline} & \textbf{91.6} \\
& \textbf{Ours} & \textbf{89.5} & & Ours & 90.8 \\
\bottomrule
\end{NiceTabular}
\end{small}
\end{center}
\end{table}

%% file: figs/fig_adc_quant_and_rescaling.tex
\begin{figure}[htb!]
\centering
\includegraphics[width=.6\linewidth]{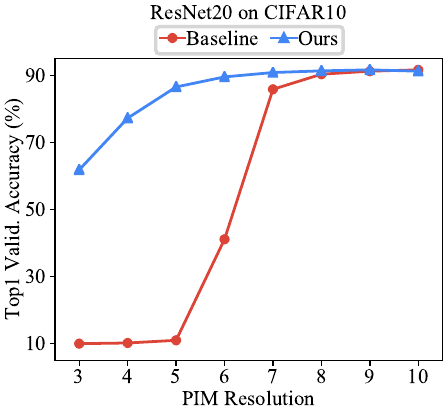}
\caption{
Comparing our method of PIM-quantization-aware training with baseline on idealized noiseless bit-serial PIM systems with different resolutions.
}
\label{fig:adc_quant_rescaling}
\end{figure}

%% file: tabs/tab_rescaling.tex
\begin{table}[t]
\caption{Ablation study of forward and backward rescaling techniques for bit-serial PIM systems with different resolutions. The accuracy results are based on ResNet20 on CIFAR10.}
\label{tab:rescaling}
\begin{center}
\begin{small}
\setlength{\tabcolsep}{3.5pt}
\renewcommand{\arraystretch}{1.0}
\scalebox{.9}{
\begin{tabular}{cccc}
\toprule
\multirow{2}{*}{$b_{\IMC}$} & \multicolumn{2}{c}{Rescaling}  & \multirow{2}{*}{Acc.} \\
& Forward & Backward & \\
\midrule
\multirow{3}{*}{3} & N & N & 10.0 \\
& N & Y & 17.1 \\
& Y & Y & 61.8 \\
\midrule
\multirow{3}{*}{4} & N & N & 61.0 \\
& N & Y & 76.7 \\
& Y & Y & 77.2 \\
\midrule
\multirow{3}{*}{5} & N & N & 10.3 \\
& N & Y & 17.5 \\
& Y & Y & 86.5 \\
\midrule
\multirow{3}{*}{6} & N & N & 10.3 \\
& N & Y & 89.1 \\
& Y & Y & 89.5 \\
\midrule
\multirow{3}{*}{7} & N & N & 88.8 \\
& N & Y & 91.0 \\
& Y & Y & 90.8 \\
\bottomrule
\end{tabular}
}
\end{small}
\end{center}
\end{table}

%% file: figs/fig_rebuttal_training_curve.tex
\begin{figure}[!htb]
\centering
\includegraphics[width=.9\linewidth]{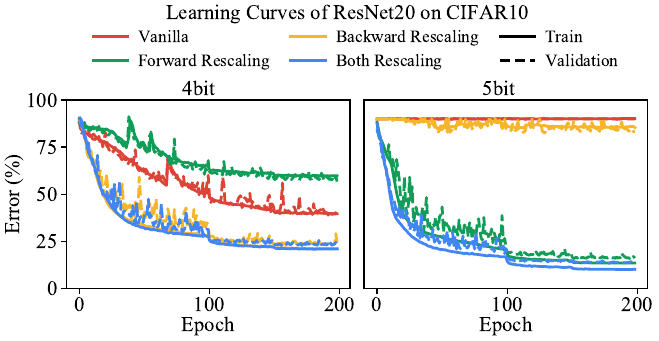}
\caption{
Learning curve comparison with ResNet20 on CIFAR10 for bit-serial PIM system.
}
\label{fig:rebuttal_training_curves}
\end{figure}

%% file: figs/fig_bn_calib.tex
\begin{figure*}[t]
\centering
\begin{subfigure}[t]{.46\linewidth}
\centering
\includegraphics[width=.8\linewidth]{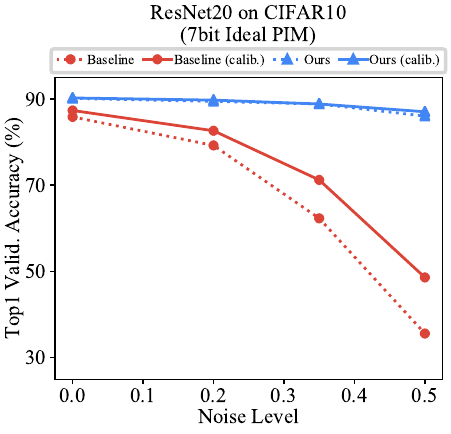}
\label{fig:bn_calib_ideal_adc}
\end{subfigure}
\hskip -4ex
\begin{subfigure}[t]{.46\linewidth}
\raisebox{0.008\linewidth}{
\centering
\includegraphics[width=.8\linewidth]{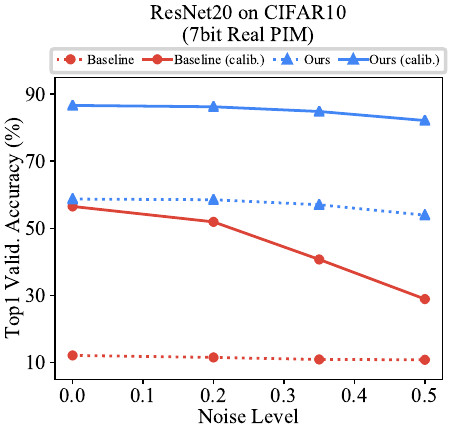}
\label{fig:bn_calib_real_adc}
}
\end{subfigure}
\caption{
Effect of BN calibration for bit-serial PIM systems with idealized and real curve quantization. We can find BN calibration helps for both our method and the baseline.
}
\label{fig:bn_calib}
\end{figure*}

%% file: tex_files/A7_bn_calib.tex
\section{More Study of BN Calibration}
\label{sec:more_bn_calib}

\input{figs/fig_more_study_on_bn_calib}
\input{tabs/tab_more_study_on_bn_calib}

Here we present more study of the effectiveness of BN calibration, and demonstrate that it is also beneficial for hardware calibrating. Specifically, we use several PIM quantization transfer curves with variation in gain and offset, as illustrated in Fig.~\ref{fig:bn_calib_fake_adc}.  The gain and offset variation is extracted from a real chip before hardware calibration. As shown in Table~\ref{tab:bn_calib_fake_adc}, directly applying these curves on a pretrained model leads to random guess results, while BN calibration is able to repair the model and recover the result to reasonable final accuracy.

%% file: figs/fig_more_study_on_bn_calib.tex
\begin{figure*}[htb!]
\centering
\begin{subfigure}[htb!]{.46\linewidth}
\centering
\includegraphics[width=.8\linewidth]{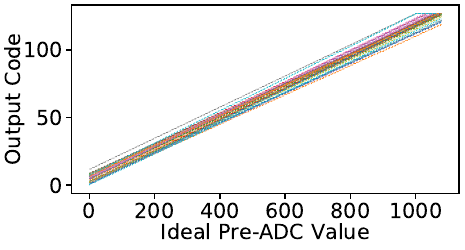}
\label{fig:bn_calib_fake_adc_72}
\end{subfigure}
\hskip -4ex
\begin{subfigure}[htb!]{.46\linewidth}
\raisebox{0.00\linewidth}{
\centering
\includegraphics[width=.8\linewidth]{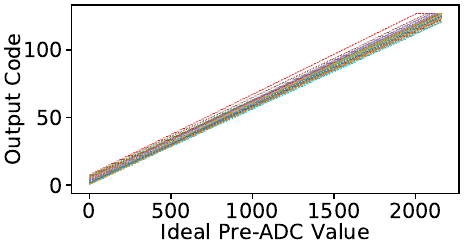}
\label{fig:bn_calib_fake_adc_144}
}
\end{subfigure}
\caption{
Generated idealized 7bit curves with gain and offset variations, where $N=72$ for (a) and $N=144$ for (b). Gains and offsets are both sampled with normal random distributions, where $\mathcal{N}_\mathrm{offset} \sim (0, 2.04)$ and $\mathcal{N}_\mathrm{gain} \sim (1, 0.024)$. The standard deviations for them are determined based on real-chip testing results.
}
\label{fig:bn_calib_fake_adc}
\end{figure*}

%% file: tabs/tab_more_study_on_bn_calib.tex
\begin{table}[htb!]
\caption{Accuracy of ResNet20 and ResNet56 on CIFAR10 with idealized bit-serial PIM-quantization with gain and offset variations, where noise or real chip curve are not involved.}
\label{tab:bn_calib_fake_adc}
\begin{center}
\begin{small}
\begin{tabular}{ccccc}
\toprule
Depth & N & Gain \& Offset Variation & BN Calib. & Acc. \\
\midrule
\multirow{6.5}{*}{20} & \multirow{3}{*}{72} & N & - & 91.2 \\
 & & Y & N & 10.0 \\
 & & Y & Y & \textbf{90.7} \\
\cmidrule{2-5}
& \multirow{3}{*}{144} & N & - & 90.8 \\
 & & Y & N & 10.0 \\
 & & Y & Y & \textbf{90.6} \\
\midrule
\multirow{6.5}{*}{56} & \multirow{3}{*}{72} & N & - & 92.2 \\
 & & Y & N & 10.0 \\
 & & Y & Y & \textbf{91.7} \\
\cmidrule{2-5}
& \multirow{3}{*}{144} & N & - & 90.3 \\
 & & Y & N & 10.1 \\
 & & Y & Y & \textbf{89.7} \\
\bottomrule
\end{tabular}
\end{small}
\end{center}
\end{table}